\documentclass[lettersize,journal]{IEEEtran} 

\usepackage[letterpaper,margin=0.75in,columnsep=0.25in]{geometry}
\usepackage[T1]{fontenc}
\usepackage[utf8]{inputenc}

\usepackage[hyphens]{url}  
\usepackage{graphicx}
\usepackage{natbib}
\usepackage{caption} 
\usepackage{amsmath}
\usepackage{amssymb}
\usepackage{amsfonts}
\usepackage{amsthm}
\usepackage{subcaption}
\usepackage{booktabs}
\usepackage{array}
\usepackage{multirow}
\usepackage{pifont}
\usepackage[table]{xcolor}
\usepackage{algorithm}
\usepackage{algorithmic}
\usepackage{xr}
\usepackage[hidelinks]{hyperref}

\IfFileExists{appendix.aux}{\externaldocument{appendix}}{}

\newtheorem{theorem}{Theorem}

\newtheorem{assumption}{Assumption}

\hypersetup{
	pdftitle={Rethinking Factor Sharing in Federated LoRA: A Rank-Aware Adaptive Approach},
	pdfauthor={Xinyi Xu, Bingnan Xiao, Shuang Qin, Gang Feng, and Tony Q.S. Quek}
}

\title{Rethinking Factor Sharing in Federated LoRA: A Rank-Aware Adaptive Approach}

\author{
Xinyi Xu\textsuperscript{1,*}\quad
Bingnan Xiao\textsuperscript{2,*}\quad
Shuang Qin\textsuperscript{1,\textdagger}\quad
Gang Feng\textsuperscript{1}\quad
Tony Q.S. Quek\textsuperscript{3}\\[0.5em]
\small \textsuperscript{1}University of Electronic Science and Technology of China\\
\small \textsuperscript{2}Fudan University\\
\small \textsuperscript{3}Singapore University of Technology and Design\\[0.25em]
\small \textsuperscript{*}Equal contribution.\quad
\textsuperscript{\textdagger}Corresponding author.\\[0.25em]
\small \texttt{xinyixu@std.uestc.edu.cn, 22110720061@m.fudan.edu.cn, blueqs@uestc.edu.cn}\\
\small \texttt{fenggang@uestc.edu.cn, tonyquek@sutd.edu.sg}
}
\date{}

\begin{document}
	
	\maketitle
	
	\begin{abstract}
		Low-rank adaptation (LoRA) represents large language model (LLM) updates with two compact matrix factors, i.e., $A$ and $B$, providing an efficient way to fine-tune large models in federated learning paradigm. 
		Inspired by the asymmetric roles of the LoRA factors, we study whether $A$ should be shared across clients while $B$ remains client-specific (Share-A/Local-B), or whether $B$ should instead be shared while $A$ remains client-specific (Share-B/Local-A).
		With a least-squares surrogate, we reveal that Share-A/Local-B requires the client-specific LoRA update matrices to use a common rank-$r$ input-side space, whereas Share-B/Local-A requires a common rank-$r$ output-side space. 
		The two strategies therefore incur different projection residuals, indicating that the preferred strategy is the one with the smaller aggregate residual across clients. 
		With this insight, we propose Federated Adaptive Factor Sharing Low-Rank Adaptation (FedAS-LoRA), which selects the sharing side before training to enhance fine-tuning performance. 
		To enable adaptive factor selection before training, we design a Rank-Aware Shared-Subspace Sufficiency (RSS) metric, which effectively assesses whether a shared rank-$r$ input subspace is sufficient for the local data distributions using representations extracted from a frozen LLM backbone. 
		Experiments across different tasks, data distributions, LoRA ranks, and participation settings confirm the effectiveness of RSS and the superior performance of FedAS-LoRA.
	\end{abstract}

	\section{Introduction}
	Large language models (LLMs) have demonstrated strong capabilities in language understanding and generation, supporting a wide range of natural language processing applications~\cite{brown2020language,touvron2023llama}. For pretrained LLMs, task-specific adaptations are needed before deployment to meet the requirements of different scenarios~\cite{houlsby2019parameter,li2021prefix,hu2022lora}. In many practical settings, the data required for fine-tuning are distributed across clients or organizations and cannot be centralized due to privacy, ownership, or regulatory constraints. Federated learning (FL) enables multiple clients to collaboratively 
	fine-tune a model without sharing their private data~\cite{mcmahan2017communication,li2020federated}. Nevertheless, full-model fine-tuning in FL incurs substantial computation, storage, and communication costs. Parameter-efficient fine-tuning provides a practical alternative by updating only a small set of additional parameters~\cite{he2022towards}. In particular, low-rank adaptation (LoRA) keeps the pretrained LLM backbone frozen and represents each model update by two trainable low-rank factors, making it well suited to resource-constrained federated adaptation.

	One critical challenge of federated LoRA lies in aggregation mismatch. For 
	an FL system with $N$ clients, let the LoRA update of client $i$ be $\Delta W_i=B_iA_i$. Independently averaging the two factors $B_i$ and $A_i$ at the server yields
	\begin{equation}
		\Big(
		\frac{1}{N}\sum_{i =1}^{N}B_i
		\Big)
		\Big(
		\frac{1}{N}\sum_{i =1}^{N}A_i
		\Big)
		\neq
		\frac{1}{N}\sum_{i =1}^{N}B_iA_i,
		\label{eq:aggregation_mismatch}
	\end{equation}
	which differs from directly averaging the client model updates. Several studies have explored methods to address this mismatch~\cite{bian2025lorafair,wang2026iflora,chen2026fedotab}. One representative line of work assigns asymmetric training and aggregation roles to the two LoRA factors. For example, FFA-LoRA fixes a common randomly initialized factor $A_0$, and only trains and aggregates $B_i$~\cite{sun2024improving}. FedSA-LoRA instead keeps both factors trainable, aggregates $A_i$, and retains $B_i$ locally for personalized adaptation~\cite{guo2025selective}. Although such methods adopt different factor-handling strategies, they hard-code
	the roles of $A$ and $B$ before training and apply the same assignment across diverse system settings. 
	\textit{It remains unclear whether this hard-coded assignment is consistently superior, prompting a reconsideration of the roles of the LoRA factors B and A in federated settings.
	}
	
	\begin{figure}[t]
		\centering
		\includegraphics[width=0.95\linewidth]{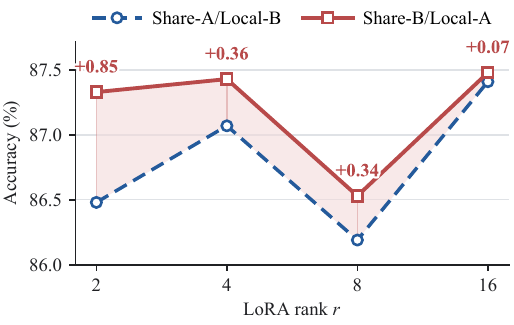}
		\caption{Accuracy gap between Share-A/Local-B and Share-B/Local-A on the MNLI-m task under label-balanced input skew. 
			The federated system includes 3 clients with full participation and 500 training rounds. 
		}
		\label{fig:motivation_mnli_rank}
	\end{figure}
	
	To examine this question, we compare Share-A/Local-B and Share-B/Local-A under a label-balanced input-skew partition. As shown in Fig.~\ref{fig:motivation_mnli_rank}, Share-B/Local-A outperforms Share-A/Local-B, and the performance gap varies with LoRA rank. This observation does not imply that Share-B/Local-A is always preferable. Instead, it means that fixing a sharing side is unsuitable for all federated LoRA settings. 
	We further analyze the structural asymmetry of the two LoRA factors through a least-squares surrogate. 
	The resultant projection characterization shows that sharing $A$ imposes a common rank-$r$ input-side representation across clients, whereas sharing $B$ imposes a common rank-$r$ output-side representation. Thus, the hard-coded sharing side assignment is not globally optimal, since
	the appropriate shared factor depends on whether a common rank-$r$ input-side or output-side space yields a smaller aggregate projection residual across clients.
	
	\begin{figure*}[t]
		\centering
		\includegraphics[width=0.85\textwidth]{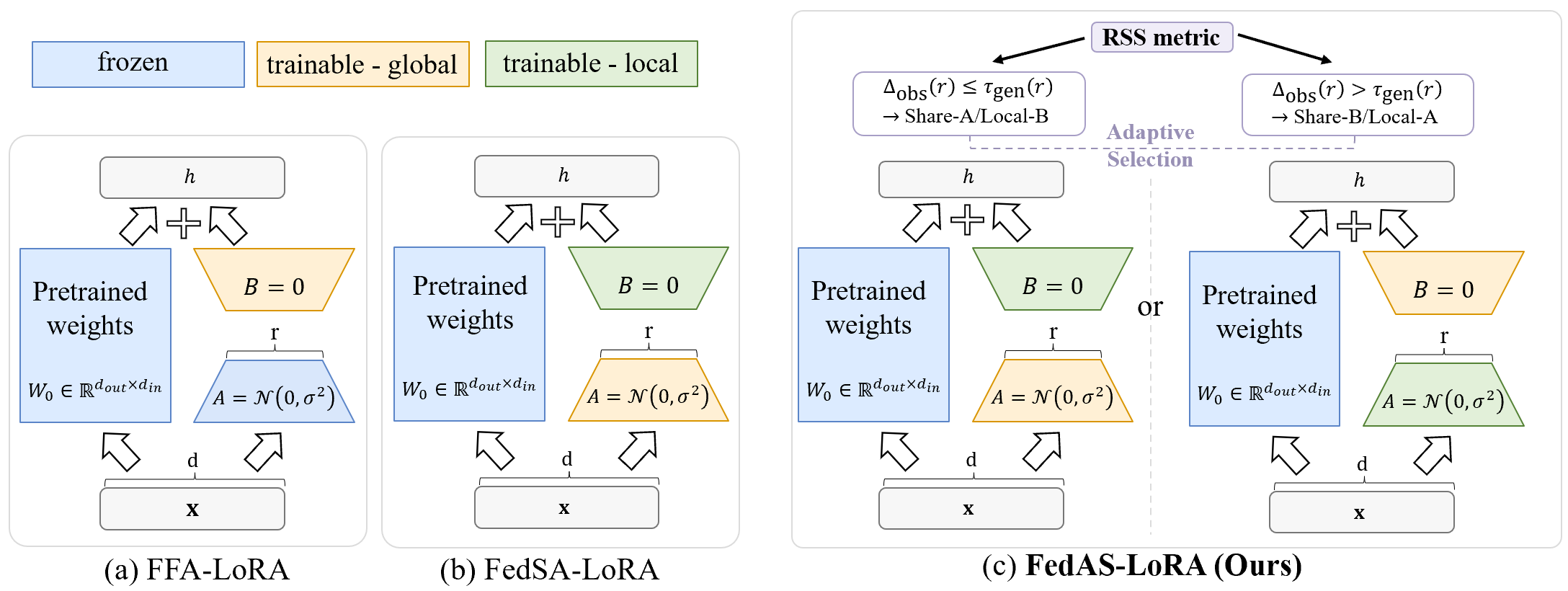}
		\caption{Illustration of (a) FFA-LoRA, (b) FedSA-LoRA, and (c) FedAS-LoRA. 
			In FFA-LoRA, $A$ is fixed after initialization, while $B$ is trainable and shared with the server for aggregation. In FedSA-LoRA, both $A$ and $B$ are trainable, while $A$ is shared and $B$ is retained locally. In FedAS-LoRA, the server uses RSS to determine the sharing strategy before training. For either strategy, both factors are updated locally; each client uploads only the shared factor and retains the other factor locally.}
		\label{fig:systemmodel}
	\end{figure*}
	
	Based on this insight, we propose Federated Adaptive Factor Sharing Low-Rank Adaptation (FedAS-LoRA) framework, as illustrated in Fig.~\ref{fig:systemmodel}. FedAS-LoRA adaptively selects the shared factor before training instead of hard-coding the same policy. To guide this selection, we design a Rank-Aware Shared-Subspace Sufficiency (RSS) metric using sequence-level representations extracted from a frozen LLM backbone. 
	RSS compares the second-order variation captured by a global rank-$r$ input subspace with that captured by the corresponding client-specific subspaces. 
	When the RSS deficit exceeds the calibrated threshold, FedAS-LoRA treats the global rank-$r$ input subspace as insufficient and selects Share-B/Local-A; otherwise, it selects Share-A/Local-B.
	We further establish that FedAS-LoRA converges to a stationary neighborhood under arbitrary client participation.
	
	Our main contributions are summarized as follows:
	\begin{itemize}
		\item 
		We provide the first projection-based comparison between Share-A/Local-B and Share-B/Local-A in federated LoRA.
		We reveal that Share-A/Local-B and Share-B/Local-A yield input-side and output-side projection residuals, respectively, which explains why fixed sharing strategies are not suitable for all federated settings.
		
		\item Based on this insight, we design an RSS metric to efficiently determine the shared factor before training. With RSS, we propose FedAS-LoRA, which adapts different federated settings to enhance fine-tuning performance.
		
		\item We prove the convergence of FedAS-LoRA under arbitrary client participation, extending the analysis beyond the full-participation setting.
		For either Share-A/Local-B or Share-B/Local-A strategy, we prove that the corresponding iterates converge to a stationary neighborhood.
		
		\item  Extensive experiments on diverse natural language tasks demonstrate the superiority of FedAS-LoRA over other methods, and validate the effectiveness of RSS-based adaptive sharing side selection.
	\end{itemize}

	\section{Related Work}
	
	\subsection{Federated Parameter-Efficient Fine-Tuning}
	Parameter-efficient fine-tuning (PEFT) reduces the computation, memory, and communication costs of adapting pretrained models in federated learning. Existing studies apply prompt tuning and other compact trainable modules while keeping the pretrained backbone frozen \cite{che2023federated,sun2024fedbpt,guo2024promptfl}. Low-rank adaptation (LoRA) represents each weight update with two low-rank factors, making it suitable for federated fine-tuning with limited client resources.
	
	Existing federated LoRA methods mainly address personalization, resource heterogeneity, and aggregation error~\cite{yang2025fedlorasurvey,koo2025towards,byun2025towards,
		shen2025pfedgpt}. FedDPA maintains global and personalized adapters to capture shared and client-specific knowledge \cite{yang2024dual}. HetLoRA supports resource-heterogeneous clients through rank self-pruning and sparsity-weighted aggregation \cite{cho2024heterogeneous}, while FlexLoRA and FLoRA aggregate LoRA modules with different ranks through reconstruction or stacking \cite{bai2024federated,wang2024flora}. FedEx-LoRA instead corrects the mismatch between factor-wise averaging and direct averaging of the resulting updates \cite{singhal2025fedex}. These methods determine shared and personalized components at the adapter level, or improve the aggregation of heterogeneous LoRA updates.

	\subsection{LoRA Factor Asymmetry and Factor-Wise Federated Aggregation}
	The two LoRA factors can exhibit different optimization and representation behaviors. LoRA-FA freezes $A$ and trains only $B$ to reduce the memory cost of fine-tuning \cite{zhang2023lorafa}, whereas LoRA+ assigns different learning rates to the two factors \cite{hayou2024loraplus}. Zhu et al.\ interpret $A$ as extracting input features and $B$ as mapping these features to the output space \cite{zhu2024asymmetry}. HydraLoRA further adopts an asymmetric architecture with a shared $A$ and multiple $B$ experts \cite{tian2024hydralora}. These studies establish factor asymmetry in centralized fine-tuning, but do not consider the division of shared and client-specific factors in federated learning.
	
	Factor-wise handling has also been explored in federated LoRA~\cite{chen2026fedotab,wang2026iflora,xu2025yoco}. FFA-LoRA fixes a common randomly initialized $A$ and trains and aggregates $B$ \cite{sun2024improving}. FedSA-LoRA trains both factors, shares $A$, and retains $B$ locally for personalization \cite{guo2025selective}. RoLoRA alternates the optimization of the two factors during federated training~\cite{chen2025rolora}, while FedRot-LoRA aligns client factors through orthogonal transformations before aggregation~\cite{zhang2026fedrotlora}. Although these methods assign different training or aggregation roles to the two factors, the sharing side is either predetermined or not explicitly considered. Which LoRA factor should be shared across clients and which should remain client-specific remains underexplored.

	\section{Preliminaries and Motivation}
	\label{sec:preliminaries}
	
	This section formulates the federated LoRA factor-sharing problem and proves that a fixed sharing side is insufficient. 
	We first define a general federated factor-sharing model covering Share-A, Share-B, and full-factor sharing, followed by a motivating example to illustrate that Share-B/Local-A can outperform Share-A/Local-B. 
	Based on a least-squares surrogate, we characterize the structural asymmetry between the two LoRA factors in federated settings.

	\subsection{Factor Sharing in Federated LoRA}
	\label{subsec:general_fed_lora_model}
	Consider a federated fine-tuning system with a server and $N$
	clients, indexed by $\mathcal{N}=\{1,2,\ldots,N\}$. Client $i$
	owns a local dataset $\mathcal{D}_i$ and optimizes a local
	objective $f_i$.
	For a target linear layer with frozen pretrained weight
	$W_0\in\mathbb{R}^{d_{\mathrm{out}}\times d_{\mathrm{in}}}$, the LoRA update of each client $i$ is $\Delta W_i=B_iA_i$. Here, $A_i\in\mathbb{R}^{r\times d_{\mathrm{in}}}$,
	$B_i\in\mathbb{R}^{d_{\mathrm{out}}\times r}$, and
	$r\ll\min\{d_{\mathrm{in}},d_{\mathrm{out}}\}$ denotes the LoRA
	rank. 
	Given an input representation
	$x\in\mathbb{R}^{d_{\mathrm{in}}}$, $A_i$ maps $x$ to an
	$r$-dimensional latent representation, while $B_i$ maps this
	representation to the output space.
	
	For federated LoRA settings, we consider a fixed shared factor type $Q\in\{A,B\}$. Let $Q_i=A_i$ when $Q=A$ and
	$Q_i=B_i$ when $Q=B$ for brevity. The corresponding personalized
	federated objective is
	\begin{equation}
		\begin{aligned}
			\min_{\{A_i,B_i\}_{i=1}^{N}}
			\quad&
			\frac{1}{N}
			\sum_{i=1}^{N}
			f_i(B_iA_i)
			\\
			\mathrm{s.t.}\quad&
			Q_i=Q_j,
			\
			\forall i,j\in\mathcal{N}.
		\end{aligned}
		\label{eq:personalized_fedlora_objective}
	\end{equation}
	Here, $f_i(B_iA_i)$ denotes the local loss evaluated
	at $W_0+B_iA_i$. When $Q=A$, the constraint shares $A$
	across clients while leaving $B_i$ client-specific, yielding
	Share-A/Local-B. When $Q=B$, it shares $B$ while leaving
	$A_i$ client-specific, yielding Share-B/Local-A.
	At training round $t$, let $Q_t$ be the shared
	factor maintained by the server and $Q_{t,i}^{e}$ its local
	copy at participating client $i$ after $e$ local updates.
	In each training round $t$, the following steps are executed:
	\begin{itemize}
		\item \textbf{Client selection and downlink transmission:}
		The server selects an arbitrary subset of $\mathcal{S}_t\subseteq\mathcal{N}$ with $S_t$ clients, and broadcasts $Q_t$, $Q \in \{A,B\}$, to clients $i\in\mathcal{S}_t$.
		
		\item \textbf{Local update:}
		Each selected client initializes the local copy of every shared factor as $Q_{t,i}^{0}=Q_t$, and performs $E$ local update steps with local learning rate $\eta_\ell$ on its local dataset $\mathcal{D}_i$. During local training, the client updates all trainable LoRA factors in its local model, while keeping non-shared factors client-local.
		
		\item \textbf{Uplink Aggregation:}
		After local training, each selected client uploads the update of the shared factor, $\Delta Q_{t,i} \!=\! Q_{t,i}^{E} \!-\! Q_t, Q\in \{A,B \}$ to the server for aggregation:
		\begin{equation}
			Q_{t+1}
			=
			Q_t
			+
			\eta_g
			\frac{1}{S_t}
			\sum_{i\in\mathcal{S}_t}
			\Delta Q_{t,i},
			\label{eq:generic_server_average_update}
		\end{equation}
		where $\eta_g$ is the server learning rate.
	\end{itemize}
	The sharing strategy determines which side of the LoRA adaptation is exposed to cross-client averaging and which side remains client-specific. 

	\subsection{Observation: Fixed Sharing Can Be Suboptimal}
	\label{subsec:motivation_study}
	
	One representative personalized federated LoRA design adopts Share-A/Local-B, i.e., sharing factor $A$ while keeping $B_i, \forall i \in \mathcal{N}$ client-specific in each training round~\cite{guo2025selective}. 
	However, it remains unclear whether sharing $A$ is always preferable to sharing $B$.
	To  {\color{black}answer} this question, we compare Share-A/Local-B and Share-B/Local-A on MNLI-m~\cite{wang2018glue} using a label-balanced input-skew partition.
	{\color{black}{To create input skew while preserving label balance, we construct a three-client partition based on the genre annotations provided by MNLI~\cite{wang2018glue}. Each client is assigned premise--hypothesis pairs from one of three genres: telephone, government, and fiction. The partition is constructed such that the clients have nearly identical distributions over the three MNLI labels. Thus, the clients share the same prediction task and similar label distributions, while their inputs originate from different textual genres. This setting introduces genre-based input heterogeneity while controlling for label skew. }}
	Fig.~\ref{fig:motivation_mnli_rank} compares their test accuracies for $r\in\{2,4,8,16\}$.
	It is noted that Share-B/Local-A achieves better performance for all rank settings, while the accuracy advantage varies with the LoRA rank $r$, indicating that the performance difference between the two sharing strategies depends on the LoRA rank $r$.

	
	Fig.~\ref{fig:motivation_mnli_rank} demonstrates that Share-A/Local-B is not uniformly optimal across different FL settings. This observation motivates the following question: 
	{\color{black}\textit{For a given LoRA rank $r$ and client data distributions, what determines which factor should be shared and which should remain client-specific?}}
	We next examine this question by analyzing the structural asymmetry between the two LoRA factors.

	\subsection{Structural Asymmetry Between LoRA Factors}
	\label{subsec:lora_structural_asymmetry}
	
	For Share-A/Local-B and  Share-B/Local-A in federated LoRA, although both strategies parameterize the update
	as $B_iA_i$, the two factors act on different sides of this
	product. $A_i$ first maps the input into an $r$-dimensional
	representation, whereas $B_i$ maps that representation to the output.
	Sharing $A$ therefore enforces a common input representation, while
	sharing $B$ enforces a common set of output directions.
	
	We clarify the difference with a single-layer least-squares
	surrogate. For each client $i \in \mathcal{N}$, let
	$\Delta_i\in\mathbb{R}^{d_{\mathrm{out}}\times d_{\mathrm{in}}}$
	denote its desired adaptation. The error of a rank-$r$ LoRA update
	$B_iA_i$ is then formulated as
	\begin{align}
		\mathcal{L}_i(B_i,A_i)
		=
		\|(\Delta_i-B_iA_i)\Sigma_i^{1/2}\|_F^2,
		\label{eq:ls_surrogate_loss}
	\end{align}
	where $\Sigma_i=\mathbb{E}[x_ix_i^\top]\succ0$ with $x_i\in\mathbb{R}^{d_{\mathrm{in}}}$ denoting the input feature. 
	Since each client uploads only the shared factor, the corresponding optimization problems can be formulated as 
	$\mathcal{E}_A=
	\min_{A,\{B_i\}_i}
	\frac{1}{N}
	\sum_{i=1}^{N}
	\|(\Delta_i-B_iA)\Sigma_i^{1/2}\|_F^2$ and $\mathcal{E}_B =
	\min_{B,\{A_i\}_i}
	\frac{1}{N}
	\sum_{i=1}^{N}
	\|(\Delta_i-BA_i)\Sigma_i^{1/2}\|_F^2$. Here, $\mathcal{E}_A$ corresponds to Share-A/Local-B and
	$\mathcal{E}_B$ to Share-B/Local-A, respectively.
	Based on the formulations of $\mathcal{E}_A$ and $\mathcal{E}_B$, we derive the following theorem to reveal the differences between Share-A/Local-B and  Share-B/Local-A.
	\begin{theorem}
		\label{thm:projection_characterization}
		Assume $\Sigma_i\succ0,\forall i \in \mathcal{N}$. Minimizing $\mathcal{E}_A$ over $\{B_i\}_i$ and $\mathcal{E}_B$ over $\{A_i\}_i$ yields the following equivalent problems over the shared factors $A$ and $B$, respectively:
		\begin{align}
			\mathcal{E}_A
			&=
			\min_A
			\frac{1}{N}
			\sum_{i=1}^{N}
			\left\|
			\Delta_i\Sigma_i^{1/2}(I-\Pi_{i,A})
			\right\|_F^2,
			\label{eq:pa_projection}\\
			\mathcal{E}_B
			&=
			\min_B
			\frac{1}{N}
			\sum_{i=1}^{N}
			\left\|
			(I-\Pi_B)\Delta_i\Sigma_i^{1/2}
			\right\|_F^2.
			\label{eq:pb_projection}
		\end{align}
		where  $\Pi_{i,A}=
		(A\Sigma_i^{1/2})^\top
		(A\Sigma_iA^\top)^\dagger
		(A\Sigma_i^{1/2})$, and $\Pi_B=
		B(B^\top B)^\dagger B^\top$.
	\end{theorem}

	From \textbf{Theorem 1}, it is noted that $\Pi_{i,A}$ keeps the part that lies in the row space of $A\Sigma_i^{1/2}$, while $I-\Pi_{i,A}$ keeps the part outside this space. 
	Similarly, $\Pi_B$ keeps the part that lies in the column space of $B$, while $I-\Pi_B$ keeps the part outside this space.
	For Share-A/Local-B, with optimal $\{B_{i,*}\}_i$, $\mathcal{E}_A$ is dominated by $\|\Delta_i\Sigma_i^{1/2}(I-\Pi_{i,A})\|_{F}^2, \forall i$, which measures the residual not covered by the shared input-side space induced by $A$.
	For Share-B/Local-A, with optimal $\{A_{i,*}\}_i$, $\|(I-\Pi_B)\Delta_i\Sigma_i^{1/2}\|_F^2$ in $\mathcal{E}_B$ denotes the parts not covered by the shared output-side space induced by $B$.
	Therefore, \textbf{Theorem~1} shows a structural asymmetry between the two sharing strategies: sharing $A$ constrains the input-side space used by all clients, whereas sharing $B$ constrains the output-side space used by all clients.
	
	The structural asymmetry between \eqref{eq:pa_projection} and \eqref{eq:pb_projection}
	explains why the same LoRA rank $r$ may lead to different preferred sharing strategies. 
	For fixed $\{\Delta_i,\Sigma_i\}_{i=1}^{N}$, $r$ determines the dimension of the shared input-side or output-side space. Increasing $r$ relaxes the rank constraint in both \eqref{eq:pa_projection} and \eqref{eq:pb_projection}, so the optimal value of neither residual can increase. Therefore, neither $\mathcal{E}_A$ nor $\mathcal{E}_B$ can increase with $r$. 
	Since the two residuals measure projection errors on different sides of $\Delta_i\Sigma_i^{1/2}$, the difference $\mathcal{E}_A-\mathcal{E}_B$ is not generally invariant to $r$. Consequently, the preferred sharing strategy under the least-squares surrogate may depend on the target LoRA rank. The sharing strategy with the smaller residual is preferred, while the other factor remains client-specific.
	This leads to a design principle: \textit{share the side on which the clients have a common rank-$r$ structure, and keep the other side local.} 
	Since both $\mathcal{E}_A$ and $\mathcal{E}_B$ depend on $\Delta_i$ and $\Sigma_i$, which are unavailable before fine-tuning, in the next section, we design a training-free metric to effectively choose the sharing side before training.

	\section{Method}
	\label{sec:method}
	
	In this section, we design a rank-aware shared-subspace sufficiency metric, RSS, to determine the shared factor before training, and prove the convergence of FedAS-LoRA under arbitrary LoRA factor selection and client participation.

	\subsection{RSS-Based Sharing-Side Selection}
	\label{subsec:srios_selection}

	The goal of RSS is to estimate the preferred sharing factor to enhance federated fine-tuning performance. \textbf{Theorem~\ref{thm:projection_characterization}} shows that this selection is determined by comparing $\mathcal{E}_A$ and $\mathcal{E}_B$ associated with common rank-$r$ input- and output-side spaces, respectively. Since both $\mathcal{E}_A$ and $\mathcal{E}_B$ depend on $\Delta_i$ and $\Sigma_i$, which are unavailable before training, RSS utilizes sequence-level representations extracted from a frozen LLM backbone to evaluate whether the shared rank-$r$ input subspace is sufficient for the local data distributions. Specifically, it compares {\color{black} the projected global-mean-centered second moment} captured by a global rank-$r$ subspace with that retained by client-specific rank-$r$ subspaces.
	
	Based on local datasets $\mathcal{D}_i$ with size $|\mathcal{D}_i|=D_i$, we define the RSS score as 
	\begin{align}\label{eq:rss}
		\mathrm{RSS}(r)
		=
		\sum_{i=1}^{N}
		p_i
		\frac{
			\mathrm{Tr}
			\left(
			U_{g,r}^{\top}S_iU_{g,r}
			\right)
		}{
			\mathrm{Tr}
			\left(
			U_{i,r}^{\top}S_iU_{i,r}
			\right)
			+
			\epsilon
		},
	\end{align}
	where $p_i = \frac{D_i}{\sum_{i=1}^N D_i}$, and $\epsilon>0$ is a small constant for numerical
	stability. 
	$S_i$ represents the centered covariance matrix of client $i$ with $U_{i,r}$ containing its top $r$ eigenvectors, and $U_{g,r}$ contains the top $r$
	eigenvectors of $S_g=\sum_{i=1}^{N}p_iS_i$. 
	$\mathrm{RSS}(r) \to 1$ indicates that the global rank-$r$ subspace preserves as much client-specific representation energy as the locally optimal rank-$r$ subspaces. 
	
	We further construct a threshold $\tau(r)$ for the RSS deficit
	$\Delta(r)=1-\mathrm{RSS}(r)$, accounting for the errors in randomly reassigning samples across clients and estimating local subspaces from finite data samples.
	$\tau(r)$ is formulated as 
	\begin{equation}
		\tau(r)=\max
		\left\{
		\tau_{\mathrm{null}}(r),
		\tau_{\mathrm{boot}}(r)
		\right\},
		\label{eq:sass_threshold}
	\end{equation}
	where $\tau_{\mathrm{null}}(r)$ is the $95$th percentile of the RSS deficits obtained by randomly reassigning samples across clients while preserving each client's sample size, and $\tau_{\mathrm{boot}}(r)$ is the empirical $95$th percentile of the representation-energy losses obtained after re-estimating each client's top-$r$ subspace from bootstrap resamples of its local data. Thus, $\Delta(r)>\tau(r)$ only when the deficit $\Delta(r)$ exceeds both thresholds, and the sharing-side strategy $\mathcal{M}(r)$ is
\begin{equation}
	\mathcal{M}(r)
	=
	\begin{cases}
		\text{Share-B/Local-A}, \mathrm{if} \ \Delta(r) \!>\! \tau(r),\\
		\text{Share-A/Local-B}, \mathrm{if} \ \Delta(r) \!\le\! \tau(r).
	\end{cases}
	\label{eq:selection_rule}
\end{equation}
From \eqref{eq:selection_rule}, {\color{black}we can see that} when $\Delta(r)>\tau(r)$, the loss in representation energy from using the global rank-$r$ subspace instead of the client-specific rank-$r$ subspaces exceeds both calibration thresholds. Since sharing $A$ requires all clients to use a common input-side subspace, FedAS-LoRA selects Share-B/Local-A in this case. When $\Delta(r)\le\tau(r)$, the observed energy loss does not exceed the calibrated threshold, which leads to Share-A/Local-B.

Before training starts, RSS is computed with representations extracted by a frozen LLM backbone. It compares the representation energy retained by the global rank-$r$ subspace with that retained by the client-specific rank-$r$ subspaces. Together with the calibrated threshold, such comparison is used in \eqref{eq:selection_rule} to decide whether $A$ is shared or kept client-specific. {\color{black}The details of the client and global subspaces, the two calibration terms, and the complete RSS workflow are provided in Appendix.}
\subsection{Convergence Analysis}
\label{subsec:convergence}
Based on the designed RSS to determine the shared LoRA factor, we establish the convergence of FedAS-LoRA under arbitrary client subset $\mathcal{S}_t$ with $|\mathcal{S}_t|=S_t$, $\forall t$. This starts with the following assumptions.
\begin{assumption}[Smoothness]
	\label{ass:smoothness}
	For each client $i \in \mathcal{N}$, the local objective $f_i$ is $L$-smooth. For $\forall$ $U_1,U_2\in\mathbb{R}^{d_{\mathrm{out}}\times d_{\mathrm{in}}}$,
	\begin{equation}
		f_i(U_2)
		\le
		f_i(U_1)
		+
		\left\langle
		\nabla f_i(U_1),
		U_2-U_1
		\right\rangle_F
		+
		\frac{L}{2}
		\|U_2-U_1\|_F^2 .
		\label{eq:smoothness_assumption}
	\end{equation}
\end{assumption}

\begin{assumption}[Unbiased and Bounded Stochastic Gradients]
	\label{ass:stochastic_gradient}
	For any client $i$, let $G_{t,i}^{e} = \nabla f_i(U_{t,i}^{e};\xi_{t,i}^{e})$ and $\bar{G}_{t,i}^{e} = \nabla f_i(U_{t,i}^{e})$, respectively. The stochastic gradient $G_{t,i}^{e}$ is unbiased and bounded with $G_{\max} > 0$, i.e.,
	\begin{equation}     \label{eq:unbiased and bounded}
		\mathbb{E}
		\left[
		G_{t,i}^{e}
		\mid
		U_{t,i}^{e}
		\right]
		= \bar{G}_{t,i}^{e} \ \mathrm{and} \ \|G_{t,i}^{e}\|_F \le G_{\max}.
	\end{equation}

\end{assumption}

\begin{assumption}[LoRA Parameter Bounds and Alignment]
	\label{ass:factor_condition}
	There exist constants $C_A,C_B>0$ and $c_A,c_B > 0$ such that
	\begin{align}
		\|A_{t,i}^{e}\|_F\le C_A
		\ \mathrm{and} \
		\|B_{t,i}^{e}\|_F\le C_B.  \label{eq:factor_norm_bound} \\
		\left\langle
		A_{t,i}^{e\top}A_{t,i}^{e},
		\bar{G}_{t,i}^{e\top}\bar{G}_{t,i}^{e}
		\right\rangle_F
		\ge
		c_A
		\|\bar{G}_{t,i}^{e}\|_F^2,
		\nonumber \\
		\left\langle
		B_{t,i}^{e}B_{t,i}^{e\top},  \bar{G}_{t,i}^{e}\bar{G}_{t,i}^{e\top}
		\right\rangle_F
		\ge
		c_B
		\|\bar{G}_{t,i}^{e}\|_F^2. \label{eq:nondegenerate}
	\end{align}
\end{assumption}
\textbf{Assumptions~\ref{ass:smoothness}} and~\textbf{\ref{ass:stochastic_gradient}} are standard in federated optimization \cite{li2020convergence,yu2019parallel,reddi2021adaptive}. 
\textbf{Assumption~\ref{ass:factor_condition}} characterizes the bilinear LoRA parameterization, as
widely used in federated LoRA analysis \cite{guo2025selective,chen2025rolora,park2025communication}. {\color{black}Specifically, the alignment inequalities in \textbf{Assumption~\ref{ass:factor_condition}} hold when the nonzero singular values of $A_{t,i}^{e}$ and $B_{t,i}^{e}$ are uniformly bounded away from zero and the realized gradient has non-vanishing projections onto their induced row and column spaces, respectively.}
The norm bounds keep the factor iterates in a bounded region. By the chain rule for $U=BA$, the gradient $\bar{G}_{t,i}^{e}$ with respect to $U$ induces the gradients $\bar{G}_{t,i}^{e}A_{t,i}^{e\top}$ and $B_{t,i}^{e\top}\bar{G}_{t,i}^{e}$ with respect to $B_{t,i}^{e}$ and $A_{t,i}^{e}$, respectively. 
Since the alignment inequalities are imposed only along the realized directions of $\bar{G}_{t,i}^{e}$, they do not impose a global full-rank condition on the low-rank factors.
The following result applies to any realized participation sequence that satisfies the gradient-mass coverage condition, as specified in the Appendix.
\begin{theorem}
	\label{thm:arbitrary_participation}
	Let $F_\star$ be a lower bound of the aggregate objective. For both Share-A/Local-B and Share-B/Local-A sharing strategies, the iterates of FedAS-LoRA satisfy
	\begin{align}
		&\frac{1}{NT}\sum_{t=0}^{T-1}\sum_{i=1}^{N}
		\mathbb{E}\!\left[\|\nabla f_i(U_{t,i})\|_F^2\right] \le
		\frac{F_0-F_\star}{\eta_{\ell} c T}
		\notag\\
		&+
		\frac{1}{\eta_{\ell} c T}\sum_{t=0}^{T-1}\left[\frac{S_t}{N}\mathcal{R}_{\mathrm{sel}}+\left(1-\frac{S_t}{N}\right)\mathcal{R}_{\mathrm{uns}}\right],
		\label{eq:convergence_bound}
	\end{align}
	where $c$ is a positive constant, $C_\star:=\max\{C_A,C_B\}$, $\mathcal{R}_{\mathrm{sel}}\!=\!
	E\eta_{\ell}^{2}C_AC_BG_{\max}^{3}
	\!+\! \frac{3LE}{2} \eta_{\ell}^{2}(C_A^{4}\!+\!C_B^{4})G_{\max}^{2} \!+\! \frac{3LE}{2} \cdot
	\eta_{\ell}^{4}C_A^{2}C_B^{2}G_{\max}^{4}
	\!+\! \frac{\eta_{\ell}}{2}G_{\max}^{2} \!+\!
	(\frac{1}{2\eta_{\ell}} \!+\! \frac{L}{2})
	(1\!+\!\eta_g)^{2}\eta_{\ell}^{2}E^{2}
	C_\star^{4}G_{\max}^{2}$, and $\mathcal{R}_{\mathrm{uns}}=(
	\frac{1}{2\rho\eta_{\ell}}+\frac{L}{2}
	)\eta_g^{2}\eta_{\ell}^{2}E^{2}
	C_\star^{4}G_{\max}^{2}$.
	The proof of Theorem~\ref{thm:arbitrary_participation} is provided in the Appendix.
\end{theorem}
\textbf{Theorem~\ref{thm:arbitrary_participation}} shows that FedAS-LoRA reaches a stationary neighborhood under arbitrary client participation. The first term on the right-hand side (RHS) of \eqref{eq:convergence_bound} decreases as $O(1/T)$, while the second term serves as the non-vanishing error induced by stochastic local updates, multiple local steps, and client sampling drift. 
\begin{table*}[t]
	\centering
	\small
	\renewcommand{\arraystretch}{1.15}
	
	\begin{tabular*}{\linewidth}{
			@{\extracolsep{\fill}}llccccccc@{}
		}
		\toprule
		& Method
		& MNLI-m
		& MNLI-mm
		& SST-2
		& QNLI
		& QQP
		& RTE
		& Avg. \\
		\midrule
		
		\multirow{5}{*}{LoRA}
		& LoRA
		& 88.36\raisebox{-0.25ex}{{ $\pm$0.08}}
		& 88.65\raisebox{-0.25ex}{{ $\pm$0.05}}
		& 95.19\raisebox{-0.25ex}{{ $\pm$0.09}}
		& 90.80\raisebox{-0.25ex}{{ $\pm$0.89}}
		& 85.42\raisebox{-0.25ex}{{ $\pm$1.27}}
		& 85.84\raisebox{-0.25ex}{{ $\pm$0.36}}
		& 89.04 \\
		
		& FFA-LoRA
		& 86.46\raisebox{-0.25ex}{{ $\pm$0.04}}
		& 87.47\raisebox{-0.25ex}{{ $\pm$0.09}}
		& 95.53\raisebox{-0.25ex}{{ $\pm$0.03}}
		& 89.24\raisebox{-0.25ex}{{ $\pm$0.84}}
		& 86.38\raisebox{-0.25ex}{{ $\pm$0.53}}
		& 87.05\raisebox{-0.25ex}{{ $\pm$0.07}}
		& 88.69 \\
		
		& FedDPA-LoRA
		& 88.77\raisebox{-0.25ex}{{ $\pm$0.06}}
		& 88.02\raisebox{-0.25ex}{{ $\pm$0.10}}
		& 96.10\raisebox{-0.25ex}{{ $\pm$0.05}}
		& 89.92\raisebox{-0.25ex}{{ $\pm$0.59}}
		& 86.60\raisebox{-0.25ex}{{ $\pm$0.81}}
		& 87.36\raisebox{-0.25ex}{{ $\pm$0.15}}
		& 89.46 \\
		
		& FedSA-LoRA
		& 89.77\raisebox{-0.25ex}{{ $\pm$0.02}}
		& 87.80\raisebox{-0.25ex}{{ $\pm$0.09}}
		& 95.72\raisebox{-0.25ex}{{ $\pm$0.03}}
		& 91.13\raisebox{-0.25ex}{{ $\pm$0.56}}
		& 86.87\raisebox{-0.25ex}{{ $\pm$0.64}}
		& 87.75\raisebox{-0.25ex}{{ $\pm$0.02}}
		& 89.84 \\
		
		& FedAS-LoRA (Ours)
		& \textbf{89.95}\raisebox{-0.25ex}{{ $\pm$0.06}}
		& \textbf{88.86}\raisebox{-0.25ex}{{ $\pm$0.49}}
		& \textbf{97.17}\raisebox{-0.25ex}{{ $\pm$0.19}}
		& \textbf{92.94}\raisebox{-0.25ex}{{ $\pm$0.39}}
		& \textbf{87.95}\raisebox{-0.25ex}{{ $\pm$0.56}}
		& \textbf{87.77}\raisebox{-0.25ex}{{ $\pm$0.04}}
		& \textbf{90.77} \\
		
		\midrule
		
		\multirow{5}{*}{rsLoRA}
		& rsLoRA
		& 84.46\raisebox{-0.25ex}{{ $\pm$0.27}}
		& 86.99\raisebox{-0.25ex}{{ $\pm$0.13}}
		& 96.49\raisebox{-0.25ex}{{ $\pm$0.09}}
		& 89.29\raisebox{-0.25ex}{{ $\pm$1.44}}
		& 86.11\raisebox{-0.25ex}{{ $\pm$0.87}}
		& 86.42\raisebox{-0.25ex}{{ $\pm$0.22}}
		& 88.29 \\
		
		& FFA-rsLoRA
		& 89.10\raisebox{-0.25ex}{{ $\pm$0.12}}
		& 87.85\raisebox{-0.25ex}{{ $\pm$0.25}}
		& 96.10\raisebox{-0.25ex}{{ $\pm$0.08}}
		& 89.13\raisebox{-0.25ex}{{ $\pm$0.58}}
		& 86.02\raisebox{-0.25ex}{{ $\pm$0.34}}
		& 86.98\raisebox{-0.25ex}{{ $\pm$0.09}}
		& 89.20 \\
		
		& FedDPA-rsLoRA
		& 89.36\raisebox{-0.25ex}{{ $\pm$0.19}}
		& 88.67\raisebox{-0.25ex}{{ $\pm$0.29}}
		& 96.79\raisebox{-0.25ex}{{ $\pm$0.10}}
		& 90.89\raisebox{-0.25ex}{{ $\pm$1.17}}
		& 86.68\raisebox{-0.25ex}{{ $\pm$0.02}}
		& 86.71\raisebox{-0.25ex}{{ $\pm$0.28}}
		& 89.85 \\
		
		& FedSA-rsLoRA
		& 90.28\raisebox{-0.25ex}{{ $\pm$0.09}}
		& 88.13\raisebox{-0.25ex}{{ $\pm$0.12}}
		& 96.75\raisebox{-0.25ex}{{ $\pm$0.06}}
		& 91.34\raisebox{-0.25ex}{{ $\pm$0.67}}
		& 86.65\raisebox{-0.25ex}{{ $\pm$0.28}}
		& 87.52\raisebox{-0.25ex}{{ $\pm$0.07}}
		& 90.12 \\
		
		& FedAS-rsLoRA (Ours)
		& \textbf{90.61}\raisebox{-0.25ex}{{ $\pm$0.10}}
		& \textbf{89.79}\raisebox{-0.25ex}{{ $\pm$0.23}}
		& \textbf{97.02}\raisebox{-0.25ex}{{ $\pm$0.07}}
		& \textbf{91.90}\raisebox{-0.25ex}{{ $\pm$0.45}}
		& \textbf{87.11}\raisebox{-0.25ex}{{ $\pm$0.11}}
		& \textbf{87.58}\raisebox{-0.25ex}{{ $\pm$0.08}}
		& \textbf{90.67} \\
		
		\midrule
		
		\multirow{5}{*}{VeRA}
		& VeRA
		& 89.57\raisebox{-0.25ex}{{ $\pm$0.02}}
		& 87.79\raisebox{-0.25ex}{{ $\pm$0.11}}
		& 94.33\raisebox{-0.25ex}{{ $\pm$0.17}}
		& 91.94\raisebox{-0.25ex}{{ $\pm$0.52}}
		& 88.67\raisebox{-0.25ex}{{ $\pm$0.48}}
		& 85.63\raisebox{-0.25ex}{{ $\pm$0.20}}
		& 89.66 \\
		
		& FFA-VeRA
		& 87.06\raisebox{-0.25ex}{{ $\pm$0.13}}
		& 86.72\raisebox{-0.25ex}{{ $\pm$0.41}}
		& 93.12\raisebox{-0.25ex}{{ $\pm$0.35}}
		& 90.59\raisebox{-0.25ex}{{ $\pm$0.11}}
		& 87.22\raisebox{-0.25ex}{{ $\pm$0.04}}
		& 85.15\raisebox{-0.25ex}{{ $\pm$0.29}}
		& 88.31 \\
		
		& FedDPA-VeRA
		& 87.98\raisebox{-0.25ex}{{ $\pm$0.46}}
		& 86.78\raisebox{-0.25ex}{{ $\pm$0.36}}
		& 94.39\raisebox{-0.25ex}{{ $\pm$0.48}}
		& 91.56\raisebox{-0.25ex}{{ $\pm$0.31}}
		& 88.71\raisebox{-0.25ex}{{ $\pm$0.35}}
		& 86.24\raisebox{-0.25ex}{{ $\pm$0.06}}
		& 89.28 \\
		
		& FedSA-VeRA
		& 89.73\raisebox{-0.25ex}{{ $\pm$0.16}}
		& 87.26\raisebox{-0.25ex}{{ $\pm$0.03}}
		& 94.95\raisebox{-0.25ex}{{ $\pm$0.42}}
		& 92.64\raisebox{-0.25ex}{{ $\pm$0.35}}
		& 89.05\raisebox{-0.25ex}{{ $\pm$0.11}}
		& \textbf{87.24}\raisebox{-0.25ex}{{ $\pm$0.02}}
		& 90.15 \\
		
		& FedAS-VeRA (Ours)
		& \textbf{90.28}\raisebox{-0.25ex}{{ $\pm$0.54}}
		& \textbf{88.38}\raisebox{-0.25ex}{{ $\pm$0.32}}
		& \textbf{95.87}\raisebox{-0.25ex}{{ $\pm$0.12}}
		& \textbf{93.08}\raisebox{-0.25ex}{{ $\pm$0.11}}
		& \textbf{89.26}\raisebox{-0.25ex}{{ $\pm$0.22}}
		& \textbf{87.22}\raisebox{-0.25ex}{{ $\pm$0.03}}
		& \textbf{90.68} \\
		
		\bottomrule
	\end{tabular*}
	\caption{Performance of different methods on the GLUE benchmark.
		MNLI-m denotes MNLI with matched test sets, and MNLI-mm denotes
		MNLI with mismatched test sets. }
	\label{tab:main_full_participation}
\end{table*}
The proof covers both sharing strategies since their local updates are identical: each selected client updates both \(A\) and \(B\) before uploading only the shared factor. 
The only difference lies in the uploaded factor update, leading to a different drift bound, i.e., \(\|Q_{t,k}^{E}-Q_t\|_F \le \eta_\ell E C_Q G_{\max}\), where $Q\in\{A,B\}$.

\section{Experiments}
\label{sec:experiments}
In this section, we systematically evaluate the performance of FedAS-LoRA on two types of tasks: natural language understanding and natural language generation.

\subsection{Experimental Setup}
\label{subsec:exp_setup}

\paragraph{Datasets and implementation.}
For natural language understanding tasks, we evaluate FedAS-LoRA with  RoBERTa-large~\cite{liu2019roberta} on the GLUE benchmark~\cite{wang2018glue}, including MNLI, SST-2, QNLI, QQP, and RTE.  
For natural language generation tasks, we evaluate FedAS-LoRA on the GSM8K dataset~\cite{cobbe2021training} with LLaMA3-8B~\cite{meta2024llama3}. 

For natural language understanding tasks, consistent with FedSA-LoRA~\cite{guo2025selective}, we use the pre-trained RoBERTa-large model with 355M parameters~\cite{liu2019roberta} from the HuggingFace Transformers library~\cite{wolf2020transformers} as the backbone. 
LoRA modules are inserted into the query and value projection matrices of each attention layer. The default LoRA rank is set to $r=8$ with the scaling factor $\alpha=16$. 
For LoRA variants, rsLoRA adopts the same training configuration except for the rank-stabilized scaling rule~\cite{kalajdzievski2023rslora}. We use SGD for all LoRA- and rsLoRA-based methods. For VeRA~\cite{kopiczko2024vera}, we set $r=256$ and use AdamW optimizer \cite{loshchilov2019adamw} with separate learning rates for the classification head and the adapted layers.
Across all methods, we use a batch size of 128, $E=10$, and $T=500$. The local learning rate $\eta_l$ is tuned from $\{0.001,0.002,0.005,0.01,0.02,0.05\}$, and the global learning rate $\eta_g$ is set to 1. 
Experiments are conducted on NVIDIA RTX 4090 GPUs. All results are reported as mean $\pm$ standard deviation over three independent runs.
See the Appendix for more experimental details.

\paragraph{Benchmarks.}
We compare FedAS-LoRA with representative federated LoRA methods and fixed factor-sharing policies. \emph{LoRA} denotes the standard federated implementation that trains and aggregates both factors $A_i$ and $B_i$. \emph{FFA-LoRA}~\cite{sun2024improving} fixes a common randomly initialized factor $A$ and only trains and aggregates $B_i$. \emph{FedDPA-LoRA}~\cite{yang2024dual} maintains global and personalized LoRA modules to capture shared and client-specific knowledge. \emph{FedSA-LoRA}~\cite{guo2025selective} trains both factors, aggregates $A_i$, and retains $B_i$ locally, i.e., Share-A/Local-B. 

\begin{table}[t]
	\centering
	\small
	\setlength{\tabcolsep}{1mm}
	\renewcommand{\arraystretch}{1.15}
	
	\begin{tabular}{@{}lccc@{}}
		\toprule
		& QNLI & SST-2 & MNLI-m \\
		\midrule
		
		LoRA
		& 89.11\raisebox{-0.25ex}{{ $\pm$0.41}}
		& 96.74\raisebox{-0.25ex}{{ $\pm$0.67}}
		& 86.87\raisebox{-0.25ex}{{ $\pm$0.35}} \\
		
		FFA-LoRA
		& 89.78\raisebox{-0.25ex}{{ $\pm$0.68}}
		& 96.51\raisebox{-0.25ex}{{ $\pm$0.37}}
		& 87.16\raisebox{-0.25ex}{{ $\pm$0.24}} \\
		
		FedDPA-LoRA
		& 90.26\raisebox{-0.25ex}{{ $\pm$0.45}}
		& 96.79\raisebox{-0.25ex}{{ $\pm$0.21}}
		& 87.90\raisebox{-0.25ex}{{ $\pm$0.20}} \\
		
		FedSA-LoRA
		& 90.34\raisebox{-0.25ex}{{ $\pm$0.67}}
		& 96.72\raisebox{-0.25ex}{{ $\pm$0.44}}
		& 88.16\raisebox{-0.25ex}{{ $\pm$0.27}} \\
		
		FedAS-LoRA (Ours)
		& \textbf{91.04}\raisebox{-0.25ex}{{ $\pm$0.34}}
		& \textbf{97.64}\raisebox{-0.25ex}{{ $\pm$0.26}}
		& \textbf{88.67}\raisebox{-0.25ex}{{ $\pm$0.19}} \\
		
		\midrule
		
		LoRA
		& 84.15\raisebox{-0.25ex}{{ $\pm$0.52}}
		& 95.79\raisebox{-0.25ex}{{ $\pm$0.41}}
		& 83.07\raisebox{-0.25ex}{{ $\pm$0.26}} \\
		
		FFA-LoRA
		& 84.48\raisebox{-0.25ex}{{ $\pm$0.63}}
		& 95.83\raisebox{-0.25ex}{{ $\pm$0.51}}
		& 84.25\raisebox{-0.25ex}{{ $\pm$0.42}} \\
		
		FedDPA-LoRA
		& 85.45\raisebox{-0.25ex}{{ $\pm$0.77}}
		& 95.38\raisebox{-0.25ex}{{ $\pm$0.59}}
		& 85.80\raisebox{-0.25ex}{{ $\pm$0.54}} \\
		
		FedSA-LoRA
		& 86.11\raisebox{-0.25ex}{{ $\pm$0.50}}
		& 96.50\raisebox{-0.25ex}{{ $\pm$0.62}}
		& 86.27\raisebox{-0.25ex}{{ $\pm$0.43}} \\
		
		FedAS-LoRA (Ours)
		& \textbf{86.64}\raisebox{-0.25ex}{{ $\pm$0.41}}
		& \textbf{97.58}\raisebox{-0.25ex}{{ $\pm$0.30}}
		& \textbf{87.52}\raisebox{-0.25ex}{{ $\pm$0.28}} \\
		
		\bottomrule
	\end{tabular}
	\caption{Performance on QNLI, SST-2, and MNLI-m with $N=10$ clients under uniform and non-uniform sampling. The upper block uses a uniform per-round sampling rate of $0.3$, while the lower block uses independent client-specific Bernoulli sampling probabilities linearly spaced from $0.1$ to $0.5$.}
	\label{tab:sampling_results}
\end{table}
\subsection{Main Results}
\label{subsec:main_results}
{\color{black}For natural language understanding tasks, consistent with FFA-LoRA \cite{sun2024improving}, we randomly split the data across three clients for federated learning. We consider a non-IID setting using a Dirichlet distribution with $\alpha=0.5$, i.e., Dir(0.5). From Table~\ref{tab:main_full_participation}, we observe that FedAS-LoRA, FedAS-rsLoRA, and FedAS-VeRA consistently outperform the compared methods in terms of average accuracy, demonstrating the effectiveness of the proposed method. FedAS-LoRA achieves an average accuracy of $90.77\%$, which is 0.93 percentage points higher than that of FedSA-LoRA. These results indicate that applying the same fixed factor-sharing policy across tasks and adaptation variants can lead to suboptimal performance, supporting our design of selecting the sharing side according to the specific federated setting.}

{\color{black}{For natural language generation, we evaluate LLaMA3-8B~\cite{meta2024llama3} on GSM8K using the HuggingFace Transformers library~\cite{wolf2020transformers}. Following FederatedScope-LLM~\cite{kuang2023federatedscope}, we split the data across three clients under an IID distribution and adopt the same optimization settings. LoRA, FFA-LoRA, and FedAS-LoRA achieve accuracies of 55.14\%, 54.51\%, and 56.28\%, respectively, with generated examples in the Appendix.}}



\begin{table}[t]
	\centering
	\small
	\setlength{\tabcolsep}{1mm}
	\renewcommand{\arraystretch}{1.15}
	
	\begin{tabular}{@{}lccc@{}}
		\toprule
		Method & IID & Dir$(1)$ & Dir$(0.5)$ \\
		\midrule
		
		LoRA
		& 90.85\raisebox{-0.25ex}{{ $\pm$0.08}}
		& 90.96\raisebox{-0.25ex}{{ $\pm$0.46}}
		& 90.80\raisebox{-0.25ex}{{ $\pm$0.89}} \\
		
		FFA-LoRA
		& 89.51\raisebox{-0.25ex}{{ $\pm$0.25}}
		& 90.39\raisebox{-0.25ex}{{ $\pm$0.42}}
		& 89.24\raisebox{-0.25ex}{{ $\pm$0.44}} \\
		
		FedSA-LoRA
		& 91.09\raisebox{-0.25ex}{{ $\pm$0.19}}
		& 90.89\raisebox{-0.25ex}{{ $\pm$0.47}}
		& 91.13\raisebox{-0.25ex}{{ $\pm$0.57}} \\
		
		FedAS-LoRA (Ours)
		& \textbf{91.12}\raisebox{-0.25ex}{{ $\pm$0.22}}
		& \textbf{91.58}\raisebox{-0.25ex}{{ $\pm$0.11}}
		& \textbf{92.94}\raisebox{-0.25ex}{{ $\pm$0.39}} \\
		
		\midrule
		
		LoRA
		& 95.27\raisebox{-0.25ex}{{ $\pm$0.02}}
		& 95.41\raisebox{-0.25ex}{{ $\pm$0.04}}
		& 95.19\raisebox{-0.25ex}{{ $\pm$0.09}} \\
		
		FFA-LoRA
		& 95.45\raisebox{-0.25ex}{{ $\pm$0.03}}
		& 95.70\raisebox{-0.25ex}{{ $\pm$0.10}}
		& 95.53\raisebox{-0.25ex}{{ $\pm$0.03}} \\
		
		FedSA-LoRA
		& 96.09\raisebox{-0.25ex}{{ $\pm$0.04}}
		& 96.56\raisebox{-0.25ex}{{ $\pm$0.04}}
		& 95.72\raisebox{-0.25ex}{{ $\pm$0.03}} \\
		
		FedAS-LoRA (Ours)
		& \textbf{96.10}\raisebox{-0.25ex}{{ $\pm$0.02}}
		& \textbf{96.79}\raisebox{-0.25ex}{{ $\pm$0.04}}
		& \textbf{97.17}\raisebox{-0.25ex}{{ $\pm$0.19}} \\
		
		\midrule
		
		LoRA
		& 88.30\raisebox{-0.25ex}{{ $\pm$0.02}}
		& 87.78\raisebox{-0.25ex}{{ $\pm$0.05}}
		& 88.36\raisebox{-0.25ex}{{ $\pm$0.08}} \\
		
		FFA-LoRA
		& 88.69\raisebox{-0.25ex}{{ $\pm$0.03}}
		& 88.90\raisebox{-0.25ex}{{ $\pm$0.06}}
		& 86.46\raisebox{-0.25ex}{{ $\pm$0.04}} \\
		
		FedSA-LoRA
		& 89.41\raisebox{-0.25ex}{{ $\pm$0.05}}
		& 89.01\raisebox{-0.25ex}{{ $\pm$0.04}}
		& 89.77\raisebox{-0.25ex}{{ $\pm$0.02}} \\
		
		FedAS-LoRA (Ours)
		& \textbf{89.43}\raisebox{-0.25ex}{{ $\pm$0.04}}
		& \textbf{89.02}\raisebox{-0.25ex}{{ $\pm$0.03}}
		& \textbf{89.95}\raisebox{-0.25ex}{{ $\pm$0.06}} \\
		
		\bottomrule
	\end{tabular}
	\caption{Performance comparison on the QNLI, SST-2, and MNLI-m
		tasks with various degrees of data heterogeneity. From top to
		bottom, the three blocks correspond to QNLI, SST-2, and MNLI-m,
		respectively. }
	\label{tab:partition_results}
\end{table}
\subsection{In-Depth Analyses}
\label{subsec:in_depth_analyses}

We proceed to utilize LoRA-based methods to conduct analyses on the natural language understanding tasks of QNLI, SST-2, and MNLI-m to assess the impact of client sampling, data heterogeneity, and LoRA rank on model performance. {\color{black}Detailed comparisons of RSS decisions with Share-A/Local-B and Share-B/Local-A strategies, together with the communication cost analysis, are provided in the Appendix.}

\subsubsection{Effect of Client Sampling}
\label{subsec:sampling_results}

We consider $10$ clients under both uniform and non-uniform sampling settings. For uniform sampling, we apply a sampling rate of 0.3 in each round. 
For non-uniform sampling, client $i$ participates independently according to a Bernoulli distribution with probability of $0.1+0.4(i-1)/(N-1)$, i.e., the participation probabilities increase linearly from $0.1$ to $0.5$. {\color{black}As shown in Table~\ref{tab:sampling_results}, FedAS-LoRA consistently outperforms FedSA-LoRA across all evaluated tasks under both uniform and non-uniform sampling. Additional scalability results with $N=50$ clients are provided in the Appendix.}


\subsubsection{Effect of Data Heterogeneity}
\label{subsubsec:data_heterogeneity}
We consider the performance of FedAS-LoRA under different data heterogeneity levels by examining IID data distributions, Dir(1), and Dir(0.5) under full client participation.
 {\color{black}Table~\ref{tab:partition_results} shows that FedAS-LoRA outperforms the compared baselines across the evaluated IID and Dirichlet partitions. Additional results under structured input-skew partitions are provided in Appendix.}



\subsubsection{Effect of LoRA Rank}
\label{subsec:rank_awareness}
We also explore the model performance over varying LoRA ranks $r\in\{2,4,8,16\}$. The corresponding results are reported in Table~\ref{tab:rank_sweep}. It is noted that FedAS-LoRA maintains competitive performance across the evaluated ranks.  {\color{black}These results indicate that no sharing factor policy is consistently preferable across different ranks, demonstrating the effectiveness of the rank-aware sharing-side selection in FedAS-LoRA.}

\begin{table}[t]
	\centering
	\small
	\setlength{\tabcolsep}{1mm}
	\renewcommand{\arraystretch}{1.15}
	
	\begin{tabular}{@{}lccc@{}}
		\toprule
		Method & QNLI & SST-2 & MNLI-m \\
		\midrule
		
		LoRA
		& 87.12\raisebox{-0.25ex}{{ $\pm$0.16}}
		& 95.41\raisebox{-0.25ex}{{ $\pm$0.19}}
		& 88.48\raisebox{-0.25ex}{{ $\pm$0.27}} \\
		
		FFA-LoRA
		& 86.45\raisebox{-0.25ex}{{ $\pm$0.83}}
		& 94.90\raisebox{-0.25ex}{{ $\pm$0.22}}
		& 88.61\raisebox{-0.25ex}{{ $\pm$0.09}} \\
		
		FedSA-LoRA
		& 89.42\raisebox{-0.25ex}{{ $\pm$0.29}}
		& 96.56\raisebox{-0.25ex}{{ $\pm$0.41}}
		& 89.61\raisebox{-0.25ex}{{ $\pm$0.23}} \\
		
		FedAS-LoRA (Ours)
		& \textbf{89.46}\raisebox{-0.25ex}{{ $\pm$0.27}}
		& \textbf{96.79}\raisebox{-0.25ex}{{ $\pm$0.21}}
		& \textbf{89.63}\raisebox{-0.25ex}{{ $\pm$0.15}} \\
		
		\midrule
		
		LoRA
		& 90.25\raisebox{-0.25ex}{{ $\pm$0.32}}
		& 95.67\raisebox{-0.25ex}{{ $\pm$0.48}}
		& 87.38\raisebox{-0.25ex}{{ $\pm$0.35}} \\
		
		FFA-LoRA
		& 91.05\raisebox{-0.25ex}{{ $\pm$0.21}}
		& 94.78\raisebox{-0.25ex}{{ $\pm$0.49}}
		& 90.62\raisebox{-0.25ex}{{ $\pm$0.46}} \\
		
		FedSA-LoRA
		& 92.01\raisebox{-0.25ex}{{ $\pm$0.35}}
		& 96.30\raisebox{-0.25ex}{{ $\pm$0.20}}
		& 90.01\raisebox{-0.25ex}{{ $\pm$0.39}} \\
		
		FedAS-LoRA (Ours)
		& \textbf{92.80}\raisebox{-0.25ex}{{ $\pm$0.23}}
		& \textbf{96.33}\raisebox{-0.25ex}{{ $\pm$0.14}}
		& \textbf{90.28}\raisebox{-0.25ex}{{ $\pm$0.25}} \\
		
		\midrule
		
		LoRA
		& 90.80\raisebox{-0.25ex}{{ $\pm$0.29}}
		& 95.19\raisebox{-0.25ex}{{ $\pm$0.09}}
		& 88.36\raisebox{-0.25ex}{{ $\pm$0.08}} \\
		
		FFA-LoRA
		& 89.24\raisebox{-0.25ex}{{ $\pm$0.24}}
		& 95.53\raisebox{-0.25ex}{{ $\pm$0.03}}
		& 86.46\raisebox{-0.25ex}{{ $\pm$0.04}} \\
		
		FedSA-LoRA
		& 91.13\raisebox{-0.25ex}{{ $\pm$0.27}}
		& 95.72\raisebox{-0.25ex}{{ $\pm$0.03}}
		& 89.77\raisebox{-0.25ex}{{ $\pm$0.02}} \\
		
		FedAS-LoRA (Ours)
		& \textbf{92.94}\raisebox{-0.25ex}{{ $\pm$0.39}}
		& \textbf{97.17}\raisebox{-0.25ex}{{ $\pm$0.19}}
		& \textbf{89.95}\raisebox{-0.25ex}{{ $\pm$0.06}} \\
		
		\midrule
		
		LoRA
		& 88.47\raisebox{-0.25ex}{{ $\pm$0.52}}
		& 94.95\raisebox{-0.25ex}{{ $\pm$0.19}}
		& 88.75\raisebox{-0.25ex}{{ $\pm$0.27}} \\
		
		FFA-LoRA
		& 88.25\raisebox{-0.25ex}{{ $\pm$0.55}}
		& 94.92\raisebox{-0.25ex}{{ $\pm$0.16}}
		& 88.23\raisebox{-0.25ex}{{ $\pm$0.13}} \\
		
		FedSA-LoRA
		& 89.12\raisebox{-0.25ex}{{ $\pm$0.39}}
		& \textbf{95.73}\raisebox{-0.25ex}{{ $\pm$0.18}}
		& 89.01\raisebox{-0.25ex}{{ $\pm$0.17}} \\
		
		FedAS-LoRA (Ours)
		& \textbf{89.56}\raisebox{-0.25ex}{{ $\pm$0.19}}
		& \textbf{95.72}\raisebox{-0.25ex}{{ $\pm$0.16}}
		& \textbf{89.73}\raisebox{-0.25ex}{{ $\pm$0.07}} \\
		
		\bottomrule
	\end{tabular}
	\caption{Test accuracy on QNLI, SST-2, and MNLI-m with different
		LoRA ranks $r$. From top to bottom, the four blocks correspond to
		$r=2$, $r=4$, $r=8$, and $r=16$, respectively. }
	\label{tab:rank_sweep}
\end{table}

\section{Conclusion}

In this paper, we studied which LoRA factor should be shared across clients and which should remain local during federated fine-tuning. Our {\color{black}investigations have } shown that Share-A/Local-B and Share-B/Local-A induce input-side and output-side projection residuals, respectively.
To address the structural asymmetry, we proposed FedAS-LoRA, which selects the LoRA sharing factor before training and keeps the other factor client-specific. 
We designed a training-free RSS metric that compares the energy captured by global and local rank-$r$ subspaces to effectively determine the sharing factor.
We established the convergence guarantees for both sharing strategies under arbitrary client participation.
Experimental results demonstrated that the preferred sharing side can change with the data distribution, adapter rank, and participation pattern, while FedAS-LoRA achieves superior performance across the evaluated settings. 

\bibliographystyle{plainnat}
\bibliography{references}

	
	\section*{Appendix A \\ Training Process of FedAS-LoRA}
	\label{app:fedas_lora_training}
	
	\textbf{Algorithm~\ref{alg:fedas_lora_training}} summarizes the training flow of FedAS-LoRA. Before federated training, the server computes RSS from the sequence-level representations extracted by the frozen LLM backbone and determines the sharing policy $\mathcal{M}(r)$ according to \eqref{eq:selection_rule}. The selected policy remains fixed throughout training. Specifically, Share-A/Local-B sets $Q=A$, whereas Share-B/Local-A sets $Q=B$.
	
	At each training round $t$, the server selects a nonempty client subset $\mathcal{S}_t\subseteq\mathcal{N}$ and sends the current shared factor $Q_t$ to the selected clients. Each selected client initializes its shared-factor copy with $Q_t$, restores its retained factor, and updates both LoRA factors for $E$ local steps. It then uploads only the shared-factor increment $\Delta Q_{t,i}=Q_{t,i}^{E}-Q_t$ and retains the updated nonshared factor locally. The server aggregates the received increments according to \eqref{eq:generic_server_average_update}, while the client-local factors of unselected clients remain unchanged.
	For clarity, \textbf{Algorithm~\ref{alg:fedas_lora_training}} is written for one LoRA-adapted layer.
	
	\begin{algorithm}[t]
		\caption{FedAS-LoRA}
		\label{alg:fedas_lora_training}
		\begin{algorithmic}[1]
			\STATE \textbf{Input:} Client datasets $\{\mathcal{D}_i\}_{i=1}^{N}$, frozen LLM backbone, rank $r$, initial LoRA factors, rounds $T$, local steps $E$, local learning rate $\eta_{\ell}$, and server learning rate $\eta_g$.
			\STATE At the server, compute $\mathrm{RSS}(r)$, $\Delta(r)=1-\mathrm{RSS}(r)$, and $\tau(r)$, and determine $\mathcal{M}(r)$ according to \eqref{eq:selection_rule}.
			\IF{$\mathcal{M}(r)=\text{Share-A/Local-B}$}
			\STATE Set $Q=A$; the server maintains $Q_t=A_t$, and client $i$ retains $B_{t,i}$ locally.
			\ELSE
			\STATE Set $Q=B$; the server maintains $Q_t=B_t$, and client $i$ retains $A_{t,i}$ locally.
			\ENDIF
			\FOR{$t=0,1,\ldots,T-1$}
			\STATE The server selects a nonempty subset $\mathcal{S}_t\subseteq\mathcal{N}$ with $S_t=|\mathcal{S}_t|$ and sends $Q_t$ to each client $i\in\mathcal{S}_t$.
			\FOR{each client $i\in\mathcal{S}_t$ in parallel}
			\STATE Set $Q_{t,i}^{0}=Q_t$ and initialize the nonshared factor from client $i$'s retained local state.
			\FOR{$e=0,1,\ldots,E-1$}
			\STATE Sample a mini-batch $\xi_{t,i}^{e}$ and set $U_{t,i}^{e}=B_{t,i}^{e}A_{t,i}^{e}$.
			\STATE Compute $G_{t,i}^{e}=\nabla f_i(U_{t,i}^{e};\xi_{t,i}^{e})$.
			\STATE Update $B_{t,i}^{e+1}=B_{t,i}^{e}-\eta_{\ell}G_{t,i}^{e}A_{t,i}^{e\top}$.
			\STATE Update $A_{t,i}^{e+1}=A_{t,i}^{e}-\eta_{\ell}B_{t,i}^{e\top}G_{t,i}^{e}$.
			\ENDFOR
			\STATE Upload $\Delta Q_{t,i}=Q_{t,i}^{E}-Q_t$; retain $B_{t,i}^{E}$ if $Q=A$, or $A_{t,i}^{E}$ if $Q=B$.
			\ENDFOR
			\STATE The server updates
			$Q_{t+1}=Q_t+\eta_g S_t^{-1}\sum_{i\in\mathcal{S}_t}\Delta Q_{t,i}$.
			\STATE Each client $j\notin\mathcal{S}_t$ keeps its client-local factor unchanged.
			\ENDFOR
			\RETURN The shared factor $Q_T$ and the retained client-local factors.
		\end{algorithmic}
	\end{algorithm}

	\section*{Appendix B \\ Design Details of the RSS Metric}
	\label{app:cls_sass_details}

	\begin{algorithm}[t]
		\caption{RSS Computation Workflow}
		\label{alg:rss_workflow}
		\begin{algorithmic}[1]
			\STATE \textbf{Input:} Client datasets
			$\{\mathcal{D}_i\}_{i=1}^{N}$, frozen LLM backbone,
			rank $r$, and $\epsilon>0$.
			\STATE Extract the sequence-level representation $h(x)$ for every sample using the frozen LLM backbone.
			\STATE Compute the global representation mean $\bar{h}$ from all client samples.
			\FOR{each client $i\in\mathcal{N}$}
			\STATE Compute $S_i$ according to
			\eqref{eq:sass_client_cov}.
			\ENDFOR
			\STATE Compute $\mathrm{RSS}(r)$ according to
			\eqref{eq:rss}, and set
			$\Delta(r)=1-\mathrm{RSS}(r)$.
			
			\FOR{each random reassignment indexed by $b$}
			\STATE Randomly reassign the samples across clients while preserving the client sizes $\{D_i\}_{i=1}^{N}$.
			\STATE Recompute the client and global rank-$r$ subspaces from the reassigned datasets.
			\STATE Compute
			$\mathrm{RSS}_{\mathrm{null}}^{(b)}(r)$ using
			\eqref{eq:rss}, and set
			$\Delta_{\mathrm{null}}^{(b)}(r)
			=1-\mathrm{RSS}_{\mathrm{null}}^{(b)}(r)$.
			\ENDFOR
			\STATE Compute $\tau_{\mathrm{null}}(r)$ according to
			\eqref{eq:sass_null_threshold}.
			
			\FOR{each bootstrap replicate indexed by $b$}
			\FOR{each client $i\in\mathcal{N}$}
			\STATE Draw $D_i$ samples with replacement from $\mathcal{D}_i$.
			\STATE Using the same global mean $\bar{h}$, recompute the client matrix and let $U_{i,r}^{(b)}$ contain its top $r$ orthonormal eigenvectors.
			\ENDFOR
			\STATE Compute $e_{\mathrm{boot}}^{(b)}(r)$ according to
			\eqref{eq:sass_boot_error}.
			\ENDFOR
			\STATE Compute $\tau_{\mathrm{boot}}(r)$ according to
			\eqref{eq:sass_boot_threshold}.
			\STATE Set
			$\tau(r)=
			\max\{\tau_{\mathrm{null}}(r),
			\tau_{\mathrm{boot}}(r)\}$.
			
			\IF{$\Delta(r)>\tau(r)$}
			\STATE Set
			$\mathcal{M}(r)=\text{Share-B/Local-A}$.
			\ELSE
			\STATE Set
			$\mathcal{M}(r)=\text{Share-A/Local-B}$.
			\ENDIF
			\RETURN $\mathrm{RSS}(r)$, $\Delta(r)$,
			$\tau(r)$, and $\mathcal{M}(r)$.
		\end{algorithmic}
	\end{algorithm}
	
	This section provides the construction of the client and global subspaces in \eqref{eq:rss}, followed by the two calibration terms in \eqref{eq:sass_threshold}.
	
	\paragraph{Client and global subspaces.}
	Let $h(x)\in\mathbb{R}^{d_h}$ denote the sequence-level representation of input $x$ extracted from the frozen LLM backbone. Given the client datasets $\{\mathcal{D}_i\}_{i=1}^{N}$ with $D_i=|\mathcal{D}_i|$, the global representation mean is
	\begin{align}
		\bar{h}
		=
		\frac{1}{\sum_{j=1}^{N}D_j}
		\sum_{j=1}^{N}
		\sum_{(x,y)\in\mathcal{D}_j}
		h(x).
	\end{align}
	For each client $i\in\mathcal{N}$, we then define
	\begin{align}
		S_i
		=
		\frac{1}{D_i}
		\sum_{(x,y)\in\mathcal{D}_i}
		\left(h(x)-\bar{h}\right)
		\left(h(x)-\bar{h}\right)^{\top}.
		\label{eq:sass_client_cov}
	\end{align}
	All clients are centered using the same global mean $\bar{h}$. Hence, $S_i$ reflects both the variation of client $i$'s representations and the difference between its representation mean and the global mean.
	
	Let $U_{i,r}\in\mathbb{R}^{d_h\times r}$ contain the top $r$ orthonormal eigenvectors of $S_i$. The global matrix is $ S_g
	= \sum_{i=1}^N p_i S_i =
	\sum_{i=1}^{N} \frac{D_i}{\sum_{i=1}^N D_i} S_i$, with $U_{g,r}\in\mathbb{R}^{d_h\times r}$ containing its top $r$ orthonormal eigenvectors.
	By the Rayleigh--Ritz principle, $\mathrm{Tr}(U_{i,r}^{\top}S_iU_{i,r})$ is the maximum representation energy that a rank-$r$ orthonormal subspace can retain for client $i$. In contrast, $\mathrm{Tr}(U_{g,r}^{\top}S_iU_{g,r})$ is the energy retained for the same client by the global rank-$r$ subspace. Their ratio in \eqref{eq:rss} therefore measures how much of client $i$'s locally attainable rank-$r$ energy is preserved by the global subspace. RSS averages these ratios across clients using $\{p_i\}_{i=1}^{N}$.
	
	\paragraph{Random-split calibration.}
	A nonzero deficit may arise even when samples are randomly assigned to clients. To measure this variation, we randomly reassign the samples while preserving the client sizes $\{D_i\}_{i=1}^{N}$. For the $b$-th reassignment, we recompute the client and global subspaces $S_i$ and $S_g$. Let $\mathrm{RSS}_{\mathrm{null}}^{(b)}(r)$ denote the resulting RSS score and define
	\begin{align}
		\Delta_{\mathrm{null}}^{(b)}(r)
		=
		1-\mathrm{RSS}_{\mathrm{null}}^{(b)}(r).
	\end{align}
	The random-split calibration term is
	\begin{align}
		\tau_{\mathrm{null}}(r)
		=
		Q_{0.95}
		\left(
		\left\{
		\Delta_{\mathrm{null}}^{(b)}(r)
		\right\}_{b}
		\right),
		\label{eq:sass_null_threshold}
	\end{align}
	where $Q_{0.95}$ denotes the empirical $95$th percentile. Thus, $\tau_{\mathrm{null}}(r)$ measures the RSS deficit that can be produced by random reassignment under the same client sizes.
	
	\paragraph{Bootstrap calibration.}
	The estimated local subspaces can also vary because each client has finitely many samples. For the $b$-th bootstrap replicate, client $i$ draws $D_i$ samples with replacement from $\mathcal{D}_i$. Using the same global mean $\bar{h}$, we recompute the client matrix and let $U_{i,r}^{(b)}$ contain its top $r$ orthonormal eigenvectors. The corresponding loss of representation energy is
	\begin{align}
		e_{\mathrm{boot}}^{(b)}(r)
		=
		\sum_{i=1}^{N}
		p_i
		\left[
		1-
		\frac{
			\mathrm{Tr}
			\left(
			U_{i,r}^{(b)\top}S_iU_{i,r}^{(b)}
			\right)
		}{
			\mathrm{Tr}
			\left(
			U_{i,r}^{\top}S_iU_{i,r}
			\right)
			+
			\epsilon
		}
		\right]_{+},
		\label{eq:sass_boot_error}
	\end{align}
	where $[z]_{+}=\max\{z,0\}$. The bootstrap calibration term is
	\begin{align}
		\tau_{\mathrm{boot}}(r)
		=
		Q_{0.95}
		\left(
		\left\{
		e_{\mathrm{boot}}^{(b)}(r)
		\right\}_{b}
		\right).
		\label{eq:sass_boot_threshold}
	\end{align}
	Here, $\tau_{\mathrm{boot}}(r)$ measures the loss caused by re-estimating the client-specific rank-$r$ subspaces from finite samples.
	
	The final threshold in \eqref{eq:sass_threshold} is the larger of $\tau_{\mathrm{null}}(r)$ and $\tau_{\mathrm{boot}}(r)$. Therefore, $\Delta(r)>\tau(r)$ only when the observed RSS deficit exceeds both the random-split variation and the finite-sample subspace error. The sharing policy is then determined by \eqref{eq:selection_rule}.
	
	\textbf{Algorithm~\ref{alg:rss_workflow}} summarizes the computation process of RSS, whose quantities are computed before federated optimization without gradients, LoRA warm-up, or trained LoRA factors.
	
	{\color{black}\paragraph{Relation to Theorem~1.}
		\textbf{Theorem~1} shows that, under the least-squares surrogate, the preferred sharing strategy is determined by comparing the aggregate input-side and output-side projection residuals, $\mathcal{E}_A$ and $\mathcal{E}_B$. Directly evaluating these residuals before training is infeasible because they depend on the client-specific desired adaptations $\Delta_i$, which are unavailable at that stage. RSS therefore does not directly estimate $\mathcal{E}_A$, $\mathcal{E}_B$, or their difference. Instead, it evaluates whether the common rank-$r$ input-side structure required by Share-A/Local-B is sufficient for the local data distributions, using sequence-level representations extracted from the frozen LLM backbone. When $\Delta(r)>\tau(r)$, the observed RSS deficit exceeds both the random-split variation and the finite-sample subspace error, indicating that the global rank-$r$ input subspace is insufficient. FedAS-LoRA therefore selects Share-B/Local-A. Otherwise, the observed deficit does not exceed the calibrated threshold, and Share-A/Local-B is selected. Thus, RSS serves as a training-free sharing-side selector motivated by Theorem~1, rather than a direct estimator of the two projection residuals. Its agreement with the empirically better fixed sharing strategy is reported in Table~\ref{tab:rss_metric_vs_share_acc_tau_gen}.}

	\section*{Appendix C \\ Proof of Theorem~\ref{thm:projection_characterization}}
	\label{app:proof_projection_characterization}
	

	For the Share-A/Local-B case, fix $A\in\mathbb{R}^{r\times d_{\mathrm{in}}}$, and let
	$Y_i=\Delta_i\Sigma_i^{1/2}$ and
	$X_i=A\Sigma_i^{1/2}$. Optimizing the client-local factor $B_i$ gives the least-squares problem
	$\min_{B_i}\|Y_i-B_iX_i\|_F^2$, and the corresponding solution is given by
	\begin{align}
		B_i^\star(A)
		&=
		Y_iX_i^\top
		(X_iX_i^\top)^\dagger
		=\Delta_i\Sigma_iA^\top
		(A\Sigma_iA^\top)^\dagger.
		\label{eq:bi_star_given_a}
	\end{align}
	With the definition of $\Pi_{i,A}$ in \textbf{Theorem~\ref{thm:projection_characterization}}, we have
	$B_i^\star(A)X_i=Y_i\Pi_{i,A}$. Therefore,
	\begin{align}
		&\min_{B_i}
		\|(\Delta_i-B_iA)\Sigma_i^{1/2}\|_F^2
		=
		\|Y_i(I-\Pi_{i,A})\|_F^2
		\notag\\
		&=
		\left\|
		\Delta_i\Sigma_i^{1/2}
		(I-\Pi_{i,A})
		\right\|_F^2.
	\end{align}
	Averaging over the clients and minimizing over the shared factor $A$ yields \eqref{eq:pa_projection}.

	For Share-B/Local-A, fix $B\in\mathbb{R}^{d_{\mathrm{out}}\times r}$ and again let
	$Y_i=\Delta_i\Sigma_i^{1/2}$. Since $\Sigma_i\succ0$, the change of variables $C_i=A_i\Sigma_i^{1/2}$ is bijective. The local problem can therefore be written as
	$\min_{C_i}\|Y_i-BC_i\|_F^2$. Taking
	$C_i^\star=B^\dagger Y_i$ gives
	\begin{align}
		A_i^\star(B)
		=
		C_i^\star\Sigma_i^{-1/2}
		=
		B^\dagger\Delta_i.
		\label{eq:ai_star_given_b}
	\end{align}
	The fitted value satisfies
	$BC_i^\star=BB^\dagger Y_i=\Pi_BY_i$, where $\Pi_B$ is the orthogonal projector onto the column space of $B$. It follows that
	\begin{align}
		&\min_{A_i}
		\|(\Delta_i-BA_i)\Sigma_i^{1/2}\|_F^2
		=
		\|(I-\Pi_B)Y_i\|_F^2
		\notag\\
		&=
		\left\|
		(I-\Pi_B)
		\Delta_i\Sigma_i^{1/2}
		\right\|_F^2.
	\end{align}
	Averaging over the clients and minimizing over the shared factor $B$ yields \eqref{eq:pb_projection}, which completes the proof.

	\section*{Appendix D \\ Proof of Theorem~\ref{thm:arbitrary_participation}}
	\label{app:convergence_proof}
	
	
	For client $i$ at local step $e$ of round $t$, let
	$U_{t,i}^{e}=B_{t,i}^{e}A_{t,i}^{e}$ and
	$U_{t,i}=U_{t,i}^{0}$. We define the aggregate objective as
	$F_t=\frac{1}{N}\sum_{i=1}^{N}f_i(U_{t,i})$ and recall that
	$C_\star=\max\{C_A,C_B\}$. By
	\textbf{Assumption~\ref{ass:stochastic_gradient}} and Jensen's inequality, we have
	\begin{align}
		\|\bar{G}_{t,i}^{e}\|_F
		\le
		\mathbb{E}
		\left[
		\|G_{t,i}^{e}\|_F
		\mid U_{t,i}^{e}
		\right]
		\le
		G_{\max}.
	\end{align}
	
	During local training, both LoRA factors are updated according to $    B_{t,i}^{e+1}
	=
	B_{t,i}^{e}
	-
	\eta_{\ell}
	G_{t,i}^{e}
	A_{t,i}^{e\top}, 
	A_{t,i}^{e+1}
	=
	A_{t,i}^{e}
	-
	\eta_{\ell}
	B_{t,i}^{e\top}
	G_{t,i}^{e}$.
	Thus, we have
	\begin{align}
		U_{t,i}^{e+1}-U_{t,i}^{e}
		&=
		-\eta_{\ell}
		B_{t,i}^{e}B_{t,i}^{e\top}G_{t,i}^{e}
		-
		\eta_{\ell}
		G_{t,i}^{e}A_{t,i}^{e\top}A_{t,i}^{e}
		\notag\\
		&+
		\eta_{\ell}^{2}
		G_{t,i}^{e}A_{t,i}^{e\top}
		B_{t,i}^{e\top}G_{t,i}^{e}.
		\label{eq:product_change}
	\end{align}
	
	Applying \textbf{Assumption~\ref{ass:smoothness}} to the increment \eqref{eq:product_change} gives
	\begin{align}
		f_i(U_{t,i}^{e+1})-f_i(U_{t,i}^{e})
		&\le
		\left\langle
		\bar{G}_{t,i}^{e},
		U_{t,i}^{e+1}-U_{t,i}^{e}
		\right\rangle_F
		\notag\\
		&+
		\frac{L}{2}
		\|U_{t,i}^{e+1}-U_{t,i}^{e}\|_F^2.
		\label{eq:smooth_local}
	\end{align}
	For the first term on the RHS of \eqref{eq:smooth_local},
	based on the unbiasedness of
	$G_{t,i}^{e}$ and \textbf{Assumption~\ref{ass:factor_condition}}, we have
	\begin{align}
		-\eta_{\ell}
		\mathbb{E}
		\left[
		\left\langle
		\bar{G}_{t,i}^{e},
		B_{t,i}^{e}B_{t,i}^{e\top}G_{t,i}^{e}
		\right\rangle_F
		\right]
		\!\le\!
		-\eta_{\ell}c_B
		\|\bar{G}_{t,i}^{e}\|_F^2,
		\notag\\
		-\eta_{\ell}
		\mathbb{E}
		\left[
		\left\langle
		\bar{G}_{t,i}^{e},
		G_{t,i}^{e}A_{t,i}^{e\top}A_{t,i}^{e}
		\right\rangle_F
		\right]
		\!\le\!
		-\eta_{\ell}c_A
		\|\bar{G}_{t,i}^{e}\|_F^2.
		\label{eq:first_order_terms_app}
	\end{align}
	For the $\eta_{\ell}^{2}
	G_{t,i}^{e}A_{t,i}^{e\top}
	B_{t,i}^{e\top}G_{t,i}^{e}$ term in \eqref{eq:product_change}, we have
	\begin{align} \label{temp 1}
		&\eta_{\ell}^{2}
		\left|
		\left\langle
		\bar{G}_{t,i}^{e},
		G_{t,i}^{e}A_{t,i}^{e\top}
		B_{t,i}^{e\top}G_{t,i}^{e}
		\right\rangle_F
		\right|
		\le
		\eta_{\ell}^{2}
		\|\bar{G}_{t,i}^{e}\|_F
		\|G_{t,i}^{e}\|_F^2
		\notag\\
		&\quad
		\|A_{t,i}^{e}\|_F
		\|B_{t,i}^{e}\|_F 
		\le
		\eta_{\ell}^{2}C_AC_BG_{\max}^{3}.
	\end{align}
	where the first inequality is due to the Cauchy--Schwarz inequality, and the second inequality comes from \textbf{Assumptions 2} and \textbf{3}.
	
	With \eqref{eq:first_order_terms_app}, \eqref{temp 1} and the Cauchy--Schwarz inequality, we obtain
	\begin{align}
		\|U_{t,i}^{e+1}-U_{t,i}^{e}\|_F^2
		&\le
		3\eta_{\ell}^{2}
		(C_A^{4}+C_B^{4})G_{\max}^{2}
		\notag\\
		&+3\eta_{\ell}^{4}
		C_A^{2}C_B^{2}G_{\max}^{4}.
		\label{eq:higher_order_bounds_app}
	\end{align}
	
	Combining the upper bounds above with \eqref{eq:smooth_local}, the second term on the RHS of \eqref{eq:smooth_local} is bounded, as given by
	\begin{align}
		&\mathbb{E}\!\!
		\left[
		f_i(U_{t,i}^{e+1})\!\!-\!\!f_i(U_{t,i}^{e})
		\!\right]
		\!\le\!
		-\eta_{\ell}(c_A\!+\!c_B)\mathbb{E}
		\!\left[
		\|\bar{G}_{t,i}^{e}\|_F^2
		\right]
		\notag\\
		&+\eta_{\ell}^{2}C_AC_BG_{\max}^{3}+\frac{3L}{2}\eta_{\ell}^{2}(C_A^{4}+C_B^{4})G_{\max}^{2}
		\nonumber\\
		&+\frac{3L}{2}\eta_{\ell}^{4}
		C_A^{2}C_B^{2}G_{\max}^{4},
		\label{eq:one_step_local_descent}
	\end{align}
	
	Summing \eqref{eq:one_step_local_descent} over
	$e=0,\ldots,E-1$ gives
	\begin{align} \label{eq temp2}
		&\mathbb{E}\!\left[f_i(U_{t,i}^{E})-f_i(U_{t,i})\right] \nonumber \\
		&\le
		-\eta_{\ell}(c_A+c_B)
		\sum_{e=0}^{E-1}\mathbb{E}\!\left[\|\bar{G}_{t,i}^{e}\|_F^2\right]+E r_{\mathrm{loc}},
	\end{align}
	where $r_{\mathrm{loc}}=
	\eta_{\ell}^{2}C_AC_BG_{\max}^{3}
	+\frac{3L}{2}\eta_{\ell}^{2}(C_A^{4}+C_B^{4})G_{\max}^{2}
	+\frac{3L}{2}\eta_{\ell}^{4}C_A^{2}C_B^{2}G_{\max}^{4}$.
	
	For the accumulated local gradients $\sum_{e=0}^{E-1}
	\mathbb{E}\!\left[\|\bar{G}_{t,i}^{e}\|_F^2\right]$ in \eqref{eq temp2}, we have
	\begin{align}
		&\sum_{e=0}^{E-1} \mathbb{E}\!\left[\|\bar{G}_{t,i}^{e}\|_F^2\right]
		=\mathbb{E}\left[\left\|\bar{G}_{t, i}^0\right\|_F^2+\sum_{e=1}^{E-1}\left\|\bar{G}_{t, i}^e\right\|_F^2\right]
		\notag\\
		&\geq \mathbb{E}\left[\left\|\bar{G}_{t, i}^0\right\|_F^2 \right]
		\geq {\kappa}{E}
		\mathbb{E}\left[\left\|\nabla f_i\left(U_{t, i}\right)\right\|_F^2 \right],
		\label{eq:local_gradient_stability_app}
	\end{align}
	where $\kappa \in [0,\frac{1}{E}]$. Substituting \eqref{eq:local_gradient_stability_app}
	into the preceding bound yields
	\begin{align}
		&\mathbb{E}\!\left[f_i(U_{t,i}^{E})-f_i(U_{t,i})\right]
		\le \notag\\
		&E r_{\mathrm{loc}}\!-\!\eta_{\ell}(c_A\!+\!c_B)\kappa E
		\mathbb{E}\!\left[\|\nabla f_i(U_{t,i})\|_F^2\right],
		\ i\in\mathcal{S}_t.
		\label{eq:local_descent_total}
	\end{align}
	
	We next bound the changes in the shared factor during local training and server aggregation.
	Telescoping the corresponding local factor updates gives
	\begin{align}
		\|A_{t,i}^{E}-A_t\|_F
		&\le \eta_{\ell}EC_BG_{\max},
		\ \text{for Share-A/Local-B},
		\notag\\
		\|B_{t,i}^{E}-B_t\|_F
		&\le \eta_{\ell}EC_AG_{\max},
		\ \text{for Share-B/Local-A}.
	\end{align}
	where both inequalities come from the triangle inequality,
	\textbf{Assumptions 2} and \textbf{3}. Consequently, either choice of the
	shared factor $Q$ satisfies
	\begin{align} \label{eq Q}
		\|Q_{t,i}^{E}-Q_t\|_F
		\le
		\eta_{\ell}EC_\star G_{\max},
		\ i\in\mathcal{S}_t.
	\end{align}
	
	Based on the server update in
	\eqref{eq:generic_server_average_update}, we obtain
	\begin{align}
		\|Q_{t+1}\!-\!Q_t\|_F
		&\!=\!
		\left\|
		\frac{\eta_g}{S_t}\!
		\sum_{k\in\mathcal{S}_t}\!(Q_{t,k}^{E}\!-\!Q_t)\!
		\right\|_F
		\!\!\leq\!\! \eta_g\eta_{\ell}EC_\star G_{\max}.
	\end{align}
	where the inequality follows from the triangle inequality and the
	upper bound of $\|Q_{t,k}^{E}-Q_t\|_F$ in \eqref{eq Q}.
	
	After server aggregation, let $U_{t+1,i}$ denote the product formed from $Q_{t+1}$ and client $i$'s retained local factor. For $i\in\mathcal{S}_t$, $U_{t+1,i}$ and $U_{t,i}^{E}$ have the same local factor and differ only in $Q_{t+1}$ and $Q_{t,i}^{E}$. For $j\notin\mathcal{S}_t$, $U_{t+1,j}$ and $U_{t,j}$ have the same local factor and differ only in $Q_{t+1}$ and $Q_t$. Based on Frobenius-norm submultiplicativity, we obtain
	\begin{align}
		&\|U_{t+1,i}\!-\!U_{t,i}^{E}\|_F
		\!\le\!
		D_{\mathrm{sel}}
		:=\!(1\!+\!\eta_g)\eta_{\ell}EC_\star^{2}G_{\max},\ i\in\mathcal{S}_t,
		\label{eq:selected_sync_change}\\
		&\|U_{t+1,j}\!-\!U_{t,j}\|_F
		\!\le\!
		D_{\mathrm{uns}}
		:=\!
		\eta_g\eta_{\ell}EC_\star^{2}G_{\max},\ j\notin\mathcal{S}_t.
		\label{eq:unselected_product_change}
	\end{align}
	where the selected-client bound uses
	$\|Q_{t+1}-Q_{t,i}^{E}\|_F
	\le\|Q_{t+1}-Q_t\|_F+\|Q_{t,i}^{E}-Q_t\|_F$,
	whereas the unselected-client bound contains only
	$\|Q_{t+1}-Q_t\|_F$. 
	
	
	For a selected client $i \in \mathcal{S}^t$, applying
	\textbf{Assumption~\ref{ass:smoothness}} between
	$U_{t,i}^{E}$ and $U_{t+1,i}$ gives
	{\small
		\begin{align}
			&\mathbb{E}\!\left[f_i(U_{t+1,i})-f_i(U_{t,i}^{E})\right]
			\!\le\!\mathbb{E}\!\left[
			\left\langle
			\nabla f_i(U_{t,i}^{E}),
			U_{t+1,i}-U_{t,i}^{E}
			\right\rangle_F
			\right]
			\notag\\
			&\!+\!\frac{L}{2}\mathbb{E}\!\left[\|U_{t+1,i}-U_{t,i}^{E}\|_F^2\!\right] \!\le\!
			\frac{\eta_{\ell}}{2}G_{\max}^{2} +\left(\frac{1}{2\eta_{\ell}}+\frac{L}{2}\right)D_{\mathrm{sel}}^{2},
			\label{eq:selected_sync_bound}
	\end{align}}
	where the second inequality follows from
	$\langle X,Y\rangle_F
	\le\frac{\eta_{\ell}}{2}\|X\|_F^2
	+\frac{1}{2\eta_{\ell}}\|Y\|_F^2$,
	\textbf{Assumption 2}, and
	\eqref{eq:selected_sync_change}.
	
	Combining \eqref{eq:selected_sync_bound} with
	\eqref{eq:local_descent_total} yields
	\begin{align}
		&\mathbb{E}\!\left[f_i(U_{t+1,i})-f_i(U_{t,i})\right] \nonumber \\
		&\le
		-\eta_{\ell}(c_A+c_B)\kappa E \mathbb{E}\!\left[\|\nabla f_i(U_{t,i})\|_F^2\right] +\mathcal{R}_{\mathrm{sel}},
		\label{eq:selected_bound_app}
	\end{align}
	where $\mathcal{R}_{\mathrm{sel}}= E\eta_{\ell}^{2}C_AC_BG_{\max}^{3}
	+\frac{3LE}{2}\eta_{\ell}^{2}(C_A^{4}+C_B^{4})G_{\max}^{2}+\frac{3LE}{2}\eta_{\ell}^{4}C_A^{2}C_B^{2}G_{\max}^{4}
	\!+\!\frac{\eta_{\ell}}{2}G_{\max}^{2} \!+\!\left(\frac{1}{2\eta_{\ell}}\!+\!\frac{L}{2}\right)(1\!+\!\eta_g)^2\eta_{\ell}^{2}E^{2}C_\star^{4}G_{\max}^{2}$.
	
	For an unselected client $j \notin \mathcal{S}^t$, although it performs no local update, its shared
	factor changes after aggregation:
	{\small
		\begin{align}
			&\mathbb{E}\!\left[f_j(U_{t+1,j})-f_j(U_{t,j})\right]
			\le \mathbb{E}\!\left[
			\left\langle
			\nabla f_j(U_{t,j}),
			U_{t+1,j}-U_{t,j}
			\right\rangle_F
			\right]
			\notag\\
			&+\!\!\frac{L}{2}
			\mathbb{E}\!\left[\|U_{t+1,j}\!-\!U_{t,j}\|_F^2\right]
			\!\!\leq\!\! \frac{\rho\eta_{\ell}}{2}\mathbb{E}\!\left[\!\|\nabla f_j(U_{t,j})\|_F^2\!\right] \!\!+\!\mathcal{R}_{\mathrm{uns}},
			\label{eq:unselected_bound_app}
	\end{align}}
	where $\mathcal{R}_{\mathrm{uns}}:=
	\left(\frac{1}{2\rho\eta_{\ell}}+\frac{L}{2}\right)
	\eta_g^{2}\eta_{\ell}^{2}E^{2}
	C_\star^{4}G_{\max}^{2}$. $\rho>0$ is the parameter in
	$\langle X,Y\rangle_F
	\le\frac{\rho\eta_{\ell}}{2}\|X\|_F^2
	+\frac{1}{2\rho\eta_{\ell}}\|Y\|_F^2$, and the last inequality
	uses \eqref{eq:unselected_product_change}.
	
	Averaging \eqref{eq:selected_bound_app} and
	\eqref{eq:unselected_bound_app} over all clients $i \in \mathcal{N}$ yields
	\begin{align}
		&\mathbb{E}[F_{t+1}-F_t]
		\le
		\frac{S_t}{N}\mathcal{R}_{\mathrm{sel}}
		+\left(1-\frac{S_t}{N}\right)\mathcal{R}_{\mathrm{uns}}
		\notag\\
		&
		-\eta_{\ell}(c_A+c_B)\kappa E \mathbb{E}\!\left[\mathcal{G}_t^{\mathrm{sel}}\right]+\frac{\rho\eta_{\ell}}{2}\mathbb{E}\!\left[\mathcal{G}_t^{\mathrm{uns}}\right],
		\label{eq:aggregate_one_round}
	\end{align}
	where $ \mathcal{G}_t^{\mathrm{sel}}
	:=
	\frac{1}{N}
	\sum_{i\in\mathcal{S}_t}
	\|\nabla f_i(U_{t,i})\|_F^2$, and
	$\mathcal{G}_t^{\mathrm{uns}}
	:=
	\frac{1}{N}
	\sum_{i\notin\mathcal{S}_t}
	\|\nabla f_i(U_{t,i})\|_F^2$.
	
	For $\mathcal{G}_t
	:=
	\mathcal{G}_t^{\mathrm{sel}}
	+\mathcal{G}_t^{\mathrm{uns}}
	=
	\frac{1}{N}
	\sum_{i=1}^{N}
	\|\nabla f_i(U_{t,i})\|_F^2$, when $\mathcal{G}_t=0$, the terms $-\eta_{\ell}(c_A+c_B)\kappa E
	\mathbb{E}\!\left[\mathcal{G}_t^{\mathrm{sel}}\right]
	+\frac{\rho\eta_{\ell}}{2}
	\mathbb{E}\!\left[\mathcal{G}_t^{\mathrm{uns}}\right]$ in
	\eqref{eq:aggregate_one_round} vanish. Otherwise, we have
	\begin{align}
		&-\eta_{\ell}(c_A+c_B)\kappa E
		\mathbb{E}\!\left[\mathcal{G}_t^{\mathrm{sel}}\right]
		+\frac{\rho\eta_{\ell}}{2}
		\mathbb{E}\!\left[\mathcal{G}_t^{\mathrm{uns}}\right]
		\label{eq:positive_descent_coefficient}\\
		&=
		-\eta_{\ell}
		\mathbb{E}\!\left[
		\left(
		(c_A\!+\!c_B)\kappa E\lambda
		\!-\!\frac{\rho}{2}(1\!-\!\lambda)
		\right)\!
		\mathcal{G}_t
		\right]
		\!\le\!
		-\eta_{\ell}c\,
		\mathbb{E}\!\left[\mathcal{G}_t\right], \notag
	\end{align}
	where $c:=
	(c_A+c_B)\kappa E\lambda
	-\frac{\rho}{2}(1-\lambda)>0$ with $\rho < \frac{2(c_A+c_B)\kappa E \lambda }{(1-\lambda)}$, and $\lambda$ should satisfy the gradient-mass coverage condition, as given by
	\begin{align} \label{mass}
		0< \lambda \leq \frac{\mathcal{G}_t^{\mathrm{sel}}}{\mathcal{G}_t}, \forall t.
	\end{align}  
Here, \eqref{mass} requires the selected clients to account for at least a fixed fraction $\lambda$ of the total squared gradient
mass in every round with $\mathcal{G}_t>0$. Equivalently,
$G_t^{\mathrm{sel}}\geq \lambda \mathcal{G}_t$ and
$G_t^{\mathrm{uns}}\leq (1-\lambda)\mathcal{G}_t$. This condition is
imposed on the realized gradient mass rather than on the
number or sampling distribution of the selected clients.
Hence, it allows arbitrary client participation, provided that
the selected subset does not capture an arbitrarily small
fraction of the current gradient mass. Under full
participation, one can set $\lambda=1$. Under partial
participation, any round-independent lower bound
$\lambda>0$ satisfying \eqref{mass} is sufficient. Together with
$c>0$, this condition ensures that the descent contributed by
the selected clients dominates the possible objective increase
of the unselected clients caused by the shared-factor update.
	

	Substituting \eqref{eq:positive_descent_coefficient} into \eqref{eq:aggregate_one_round} and
	rearranging gives
	\begin{align}
		\eta_{\ell}c\,
		\mathbb{E}\!\left[\mathcal{G}_t\right]
		\!\le\!
		\mathbb{E}[F_t\!-\!F_{t+1}]
		\!+\!\frac{S_t}{N}\mathcal{R}_{\mathrm{sel}}
		\!+\!\left(1-\frac{S_t}{N}\right)\mathcal{R}_{\mathrm{uns}}.
		\label{eq:one_round_rearranged}
	\end{align}
	
	Summing \eqref{eq:one_round_rearranged} over
	$t=0,\ldots,T-1$ and dividing by $\eta_l c T$, we obtain
	\begin{align}
		&\frac{1}{NT}
		\sum_{t=0}^{T-1}\sum_{i=1}^{N}
		\mathbb{E}\!\left[
		\|\nabla f_i(U_{t,i})\|_F^2
		\right] \le
		\frac{F_0-F_\star}{\eta_{\ell}cT} \nonumber \\
		&+\!\frac{1}{\eta_{\ell}cT}
		\sum_{t=0}^{T-1}
		\frac{S_t}{N}\mathcal{R}_{\mathrm{sel}}
		\!+\!\frac{1}{\eta_{\ell}cT}\sum_{t=0}^{T-1}
		\left(1\!-\!\frac{S_t}{N}\right)
		\mathcal{R}_{\mathrm{uns}},
		\label{eq:final_convergence_bound_app}
	\end{align}
	which is exactly \eqref{eq:convergence_bound} and completes the
	proof.
	\begin{figure}[!t]
	    \centering
	    \includegraphics[width=\linewidth]{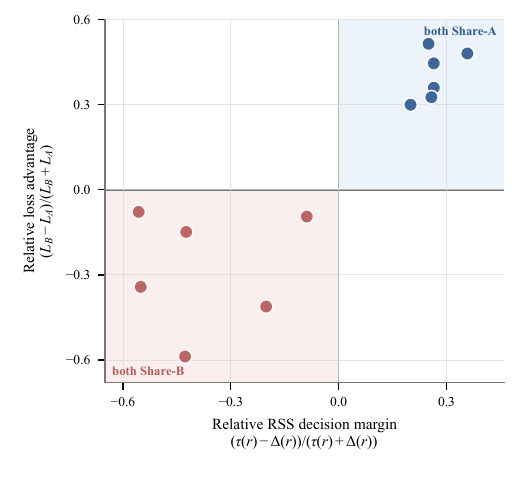}
	    \caption{ Alignment Between RSS Decisions and Training-Loss Preference}
	    \label{fig:rss_training_loss_alignment}
	\end{figure}
	
    \section*{Appendix E \\ Full Experiment Results}    
	\subsection*{Hyperparameters}
	Tables \ref{tab:lora_learning_rates} and \ref{tab:rslora_learning_rates} show the learning rates used for LoRA-based methods and rsLoRA-based methods, respectively. For the VeRA-based methods, we chose the AdamW optimizer and used separate learning rates for the classification head and the adapted layers as used in VeRA. The learning rates used for VeRA-based methods are shown in Table \ref{tab:vera_learning_rates}.
	\begin{table}[b]
		\centering
		\scriptsize
		\setlength{\tabcolsep}{1.5pt}
		\renewcommand{\arraystretch}{1.08}
		\begin{tabular*}{0.98\columnwidth}{@{\extracolsep{\fill}}lcccccc@{}}
			\toprule
			Method & MNLI-m & MNLI-mm & SST-2 & QNLI & QQP & RTE \\
			\midrule
			LoRA
			& 1E-2 & 1E-2 & 2E-2 & 1E-2 & 1E-2 & 1E-2 \\
			
			FFA-LoRA
			& 5E-2 & 5E-2 & 5E-2 & 2E-2 & 5E-2 & 2E-2 \\
			
			FedDPA-LoRA
			& 1E-2 & 1E-2 & 1E-2 & 5E-2 & 5E-2 & 1E-2 \\
			
			FedSA-LoRA
			& 2E-2 & 2E-2 & 1E-2 & 5E-3 & 2E-2 & 1E-2 \\
			
			FedAS-LoRA
			& 2E-2 & 2E-2 & 1E-2 & 5E-3 & 2E-2 & 1E-2 \\
			\bottomrule
		\end{tabular*}
		\caption{Learning rates used for LoRA-based methods on the GLUE benchmark.}
		\label{tab:lora_learning_rates}
	\end{table}
	\begin{table}[h]
		\centering
		\small
		\setlength{\tabcolsep}{1.5pt}
		\renewcommand{\arraystretch}{1.08}
		\begin{tabular*}{0.98\columnwidth}{@{\extracolsep{\fill}}lcccccc@{}}
			\toprule
			Method & MNLI-m & MNLI-mm & SST-2 & QNLI & QQP & RTE \\
			\midrule
			rsLoRA
			& 5E-3 & 5E-3 & 1E-2 & 2E-3 & 5E-3 & 2E-3 \\
			
			FFA-rsLoRA
			& 2E-2 & 2E-2 & 2E-2 & 1E-2 & 2E-2 & 1E-2 \\
			
			FedDPA-rsLoRA
			& 5E-3 & 5E-3 & 1E-2 & 1E-3 & 5E-3 & 1E-2 \\
			
			FedSA-rsLoRA
			& 5E-3 & 5E-3 & 5E-3 & 1E-3 & 2E-3 & 2E-3 \\
			
			FedAS-rsLoRA
			& 5E-3 & 5E-3 & 5E-3 & 1E-3 & 2E-3 & 2E-3 \\
			\bottomrule
		\end{tabular*}
		\caption{Learning rates used for rsLoRA-based methods on the GLUE benchmark.}
		\label{tab:rslora_learning_rates}
	\end{table}
	\begin{table}[t]
		\centering
		\small
		\setlength{\tabcolsep}{1.0pt}
		\renewcommand{\arraystretch}{1.08}
		\begin{tabular*}{0.98\columnwidth}{@{\extracolsep{\fill}}llcccccc@{}}
			\toprule
			Method & Position & \shortstack{MNLI-\\m} & \shortstack{MNLI-\\mm} & SST-2 & QNLI & QQP & RTE \\
			\midrule
			
			\multirow{2}{*}{VeRA}
			& VeRA & 1E-2 & 1E-2 & 2E-2 & 2E-3 & 2E-3 & 1E-2 \\
			& Head & 6E-3 & 6E-3 & 2E-3 & 3E-4 & 3E-4 & 2E-4 \\
			\midrule
			
			\multirow{2}{*}{FFA-VeRA}
			& VeRA & 2E-2 & 2E-2 & 1E-2 & 1E-2 & 1E-2 & 1E-2 \\
			& Head & 2E-3 & 2E-3 & 6E-3 & 2E-4 & 6E-3 & 2E-4 \\
			\midrule
			
			\multirow{2}{*}{FedDPA-VeRA}
			& VeRA & 1E-2 & 1E-2 & 1E-2 & 2E-3 & 2E-2 & 1E-2 \\
			& Head & 6E-3 & 6E-3 & 6E-3 & 3E-4 & 2E-3 & 2E-4 \\
			\midrule
			
			\multirow{2}{*}{FedSA-VeRA}
			& VeRA & 2E-3 & 2E-3 & 1E-2 & 1E-2 & 2E-3 & 1E-2 \\
			& Head & 3E-5 & 3E-5 & 3E-4 & 3E-4 & 3E-4 & 1E-4 \\
			\midrule
			
			\multirow{2}{*}{FedAS-VeRA}
			& VeRA & 2E-3 & 2E-3 & 1E-2 & 1E-2 & 2E-3 & 1E-2 \\
			& Head & 3E-5 & 3E-5 & 3E-4 & 3E-4 & 3E-4 & 1E-4 \\
			\bottomrule
		\end{tabular*}
		\caption{Learning rates used for VeRA-based methods on the GLUE benchmark.}
		\label{tab:vera_learning_rates}
	\end{table}
	
	\subsection*{Evaluation of RSS-Based Sharing-Side Selection}
	
	{\color{black}\paragraph{Alignment with Test-Accuracy Preference.}Table~\ref{tab:rss_metric_vs_share_acc_tau_gen} provides the full comparison between the sharing-side decisions made by RSS and the empirical test accuracies of Share-A/Local-B and Share-B/Local-A. The experiments cover six evaluation sets, multiple client partitions, and LoRA ranks $r\in\{2,4,8,16\}$. For each setting, RSS is computed before federated training, while the empirical winner is determined by comparing the test accuracies obtained at the best validation rounds of the two sharing strategies.
		
		Across the 38 evaluated settings, the RSS decision agrees with the empirically better sharing strategy in all 33 cases with a strict accuracy difference. The remaining five cases are marked as empirical ties in Table~\ref{tab:rss_metric_vs_share_acc_tau_gen}. The preferred sharing side also changes with the data partition and LoRA rank. For example, QNLI and MNLI favor Share-A/Local-B under the evaluated IID settings, whereas Share-B/Local-A is preferred in most of their Dirichlet settings. For MNLI under Dirichlet $\alpha=1$, the preferred strategy further changes from Share-B/Local-A at $r=4$ to Share-A/Local-B at $r=8$. Similarly, SST-2 favors Share-A/Local-B under IID partitions but mainly favors Share-B/Local-A under the non-IID partitions. These results support the use of a partition- and rank-aware selection rule instead of a fixed sharing policy. RSS should nevertheless be interpreted as a sharing-side selector rather than an estimator of the final accuracy gap between the two strategies.}

        {\color{black}\paragraph{Alignment with the training objective.}
Table~\ref{tab:rss_metric_vs_share_acc_tau_gen} compares the RSS
decision with the final test-accuracy preference of the two
fixed sharing strategies. We further examine whether the
decision direction of RSS agrees with their relative behavior
under the actual training objective. For visual illustration,
we randomly sample several evaluated settings and compare
their RSS decision margins with the corresponding training
losses of Share-A/Local-B and Share-B/Local-A.

For rank $r$, we define the relative RSS decision margin as
\begin{align}
    m_{\mathrm{RSS}}(r)
    =
    \frac{
        \tau(r)-\Delta(r)
    }{
        \tau(r)+\Delta(r)
    }.
    \label{eq:rss_relative_margin}
\end{align}
Since RSS selects Share-A/Local-B when
$\Delta(r)\leq\tau(r)$, a positive
$m_{\mathrm{RSS}}(r)$ favors Share-A/Local-B, whereas a
negative value favors Share-B/Local-A. We also define the
relative training-loss advantage as
\begin{align}
    m_{\mathrm{loss}}
    =
    \frac{
        L_B-L_A
    }{
        L_B+L_A
    },
    \label{eq:loss_relative_margin}
\end{align}
where $L_A$ and $L_B$ denote the training losses obtained
by Share-A/Local-B and Share-B/Local-A, respectively.
Accordingly, $m_{\mathrm{loss}}>0$ means that
Share-A/Local-B attains a lower training loss, while
$m_{\mathrm{loss}}<0$ means that Share-B/Local-A attains
a lower training loss.

As shown in Fig.~\ref{fig:rss_training_loss_alignment}, all
sampled cases lie in either the upper-right or lower-left
quadrant. Thus, the sharing side selected by RSS also attains
the lower training loss in these cases. This result provides
additional evidence that the RSS decision is consistent with
the optimization behavior of the two sharing strategies under
the actual task objective.

This directional agreement provides empirical support for
the residual-based selection principle in Theorem~1. The
theorem shows that the preferred sharing side is determined
by which shared-subspace constraint yields the smaller
aggregate projection residual. RSS translates this principle
into a training-free decision rule by assessing whether the
common rank-$r$ input-side subspace required by
Share-A/Local-B is sufficient for the client representations.
When this subspace is sufficient, RSS favors
Share-A/Local-B; otherwise, it favors Share-B/Local-A.
The consistent training-loss preference therefore indicates
that RSS identifies the sharing constraint that better fits the
client data under the actual task objective. This result
supports RSS as a practical proxy for the residual-based
sharing-side preference characterized in Theorem~\ref{tab:rss_metric_vs_share_acc_tau_gen}.}

	\begin{table*}[!t]
		\centering
		\scriptsize
		\setlength{\tabcolsep}{2.5pt}
		\renewcommand{\arraystretch}{1.08}
		\resizebox{0.93\textwidth}{!}{%
			\begin{tabular}{@{}llcccccccc@{}}
				\toprule
				\multirow{2}{*}{Dataset}
				& \multirow{2}{*}{Split}
				& \multirow{2}{*}{$r$}
				& \multicolumn{3}{c}{RSS metric}
				& \multicolumn{3}{c}{Accuracy (\%)}
				& \multirow{2}{*}{Alignment} \\
				\cmidrule(lr){4-6}\cmidrule(lr){7-9}
				& & & $\Delta(r)$ & $\tau (r)$
				& Decision & Share-A & Share-B & Winner & \\
				\midrule
				\multirow{8}{*}{\textbf{QNLI}} & \multirow{2}{*}{IID} & 4 & $2.46413\!\times\!10^{-5}$ & $3.91172\!\times\!10^{-5}$ & Share-A & \textbf{92.26} & 90.18 & \textbf{Share-A} & Match \\
				&  & 8 & $4.28222\!\times\!10^{-5}$ & $7.07299\!\times\!10^{-5}$ & Share-A & \textbf{91.12} & 90.66 & \textbf{Share-A} & Match \\
				& \multirow{4}{*}{Dirichlet $\alpha=0.5$} & 2 & $3.01960\!\times\!10^{-5}$ & $3.46345\!\times\!10^{-5}$ & Share-A & \textbf{89.46} & 88.84 & \textbf{Share-A} & Match \\
				&  & 4 & $5.89316\!\times\!10^{-5}$ & $3.80830\!\times\!10^{-5}$ & Share-B & 92.01 & \textbf{92.80} & \textbf{Share-B} & Match \\
				&  & 8 & $7.37366\!\times\!10^{-5}$ & $6.79052\!\times\!10^{-5}$ & Share-B & 91.13 & \textbf{92.94} & \textbf{Share-B} & Match \\
				&  & 16 & $1.33945\!\times\!10^{-4}$ & $1.31365\!\times\!10^{-4}$ & Share-B & 89.12 & \textbf{89.56} & \textbf{Share-B} & Match \\
				& \multirow{2}{*}{Dirichlet $\alpha=1$} & 4 & $5.90747\!\times\!10^{-5}$ & $3.92434\!\times\!10^{-5}$ & Share-B & 93.28 & \textbf{94.69} & \textbf{Share-B} & Match \\
				&  & 8 & $8.31570\!\times\!10^{-5}$ & $6.96169\!\times\!10^{-5}$ & Share-B & 90.89 & \textbf{91.58} & \textbf{Share-B} & Match \\
				\midrule
				\multirow{2}{*}{\textbf{QQP}} & \multirow{2}{*}{Dirichlet $\alpha=0.5$} & 4 & $2.11905\!\times\!10^{-5}$ & $1.28376\!\times\!10^{-5}$ & Share-B & 85.68 & \textbf{86.70} & \textbf{Share-B} & Match \\
				&  & 8 & $4.14507\!\times\!10^{-5}$ & $1.67622\!\times\!10^{-5}$ & Share-B & 86.87 & \textbf{87.95} & \textbf{Share-B} & Match \\
				\midrule
				\multirow{8}{*}{\textbf{SST-2}} & \multirow{2}{*}{IID} & 4 & $3.89196\!\times\!10^{-4}$ & $5.07871\!\times\!10^{-4}$ & Share-A & \textbf{96.65} & 96.14 & \textbf{Share-A} & Match \\
				&  & 8 & $4.35904\!\times\!10^{-4}$ & $6.02930\!\times\!10^{-4}$ & Share-A & \textbf{96.10} & 95.87 & \textbf{Share-A} & Match \\
				& \multirow{4}{*}{Dirichlet $\alpha=0.5$} & 2 & $1.45398\!\times\!10^{-3}$ & $5.83336\!\times\!10^{-4}$ & Share-B & 96.56 & \textbf{96.79} & \textbf{Share-B} & Match \\
				&  & 4 & $1.84450\!\times\!10^{-3}$ & $5.25775\!\times\!10^{-4}$ & Share-B & \textbf{96.33} & \textbf{96.33} & \textbf{Tie} & Actual tie \\
				&  & 8 & $2.20251\!\times\!10^{-3}$ & $6.38468\!\times\!10^{-4}$ & Share-B & 95.72 & \textbf{97.17} & \textbf{Share-B} & Match \\
				&  & 16 & $3.63222\!\times\!10^{-3}$ & $1.01451\!\times\!10^{-3}$ & Share-B & \textbf{95.73} & 95.72 & \textbf{Tie} & Actual tie \\
				& \multirow{2}{*}{Dirichlet $\alpha=1$} & 4 & $1.18446\!\times\!10^{-3}$ & $5.79445\!\times\!10^{-4}$ & Share-B & \textbf{97.02} & \textbf{97.02} & \textbf{Tie} & Actual tie \\
				&  & 8 & $1.43308\!\times\!10^{-3}$ & $6.62178\!\times\!10^{-4}$ & Share-B & 96.56 & \textbf{96.79} & \textbf{Share-B} & Match \\
				\midrule
				\multirow{8}{*}{\textbf{MNLI-m}} & \multirow{2}{*}{IID} & 4 & $5.98420\!\times\!10^{-6}$ & $1.02906\!\times\!10^{-5}$ & Share-A & \textbf{87.06} & 86.11 & \textbf{Share-A} & Match \\
				&  & 8 & $2.91890\!\times\!10^{-5}$ & $4.13822\!\times\!10^{-5}$ & Share-A & \textbf{89.43} & 89.06 & \textbf{Share-A} & Match \\
				& \multirow{4}{*}{Dirichlet $\alpha=0.5$} & 2 & $1.03136\!\times\!10^{-4}$ & $1.11881\!\times\!10^{-5}$ & Share-B & 89.60 & \textbf{89.63} & \textbf{Share-B} & Match \\
				&  & 4 & $4.62360\!\times\!10^{-5}$ & $9.96230\!\times\!10^{-6}$ & Share-B & 90.04 & \textbf{90.28} & \textbf{Share-B} & Match \\
				&  & 8 & $9.33286\!\times\!10^{-5}$ & $3.98546\!\times\!10^{-5}$ & Share-B & 89.75 & \textbf{89.95} & \textbf{Share-B} & Match \\
				&  & 16 & $8.07826\!\times\!10^{-5}$ & $3.45278\!\times\!10^{-5}$ & Share-B & 89.04 & \textbf{89.73} & \textbf{Share-B} & Match \\
				& \multirow{2}{*}{Dirichlet $\alpha=1$} & 4 & $1.37597\!\times\!10^{-5}$ & $1.02455\!\times\!10^{-5}$ & Share-B & 89.50 & \textbf{89.62} & \textbf{Share-B} & Match \\
				&  & 8 & $3.26219\!\times\!10^{-5}$ & $3.80408\!\times\!10^{-5}$ & Share-A & \textbf{89.02} & 88.37 & \textbf{Share-A} & Match \\
				\midrule
				\multirow{6}{*}{\textbf{MNLI-mm}} & \multirow{2}{*}{IID} & 4 & $5.98420\!\times\!10^{-6}$ & $1.02906\!\times\!10^{-5}$ & Share-A & \textbf{86.68} & 85.23 & \textbf{Share-A} & Match \\
				&  & 8 & $2.91890\!\times\!10^{-5}$ & $4.13822\!\times\!10^{-5}$ & Share-A & \textbf{88.07} & 87.91 & \textbf{Share-A} & Match \\
				& \multirow{2}{*}{Dirichlet $\alpha=0.5$} & 4 & $4.62360\!\times\!10^{-5}$ & $9.96230\!\times\!10^{-6}$ & Share-B & 88.47 & \textbf{89.97} & \textbf{Share-B} & Match \\
				&  & 8 & $9.33286\!\times\!10^{-5}$ & $3.98546\!\times\!10^{-5}$ & Share-B & 87.82 & \textbf{88.86} & \textbf{Share-B} & Match \\
				& \multirow{2}{*}{Dirichlet $\alpha=1$} & 4 & $1.37597\!\times\!10^{-5}$ & $1.02455\!\times\!10^{-5}$ & Share-B & 87.14 & \textbf{88.79} & \textbf{Share-B} & Match \\
				&  & 8 & $3.26219\!\times\!10^{-5}$ & $3.80408\!\times\!10^{-5}$ & Share-A & \textbf{90.44} & 89.97 & \textbf{Share-A} & Match \\
				\midrule
				\multirow{6}{*}{\textbf{RTE}} & \multirow{2}{*}{IID} & 4 & $2.07749\!\times\!10^{-3}$ & $3.52261\!\times\!10^{-3}$ & Share-A & \textbf{88.14} & 85.82 & \textbf{Share-A} & Match \\
				&  & 8 & $3.99091\!\times\!10^{-3}$ & $5.98761\!\times\!10^{-3}$ & Share-A & \textbf{87.50} & \textbf{87.50} & \textbf{Tie} & Actual tie \\
				& \multirow{2}{*}{Dirichlet $\alpha=0.5$} & 4 & $1.81187\!\times\!10^{-3}$ & $3.83366\!\times\!10^{-3}$ & Share-A & \textbf{88.71} & \textbf{88.71} & \textbf{Tie} & Actual tie \\
				&  & 8 & $3.63808\!\times\!10^{-3}$ & $5.98369\!\times\!10^{-3}$ & Share-A & \textbf{87.77} & 86.34 & \textbf{Share-A} & Match \\
				& \multirow{2}{*}{Dirichlet $\alpha=1$} & 4 & $1.82618\!\times\!10^{-3}$ & $3.55810\!\times\!10^{-3}$ & Share-A & \textbf{87.19} & 86.47 & \textbf{Share-A} & Match \\
				&  & 8 & $3.45208\!\times\!10^{-3}$ & $5.75546\!\times\!10^{-3}$ & Share-A & \textbf{86.26} & 85.35 & \textbf{Share-A} & Match \\
				\bottomrule
			\end{tabular}%
		}
		\caption{RSS metric values, threshold decisions, and Share-A/Local-B versus Share-B/Local-A accuracies under the evaluated three-client settings.}
		\label{tab:rss_metric_vs_share_acc_tau_gen}
	\end{table*}

	\subsection*{Scalability Analysis under Uniform and Non-Uniform Client Sampling}
	{\color{black}To further evaluate scalability, we increase the number of clients to $N=50$ under uniform and non-uniform sampling. As shown in Table~\ref{tab:sampling_results_n50}, FedAS-LoRA achieves the highest accuracy on QNLI, SST-2, and MNLI-m under both sampling settings. Compared with FedSA-LoRA, its average accuracy is higher by 0.85 and 0.76 percentage points under uniform and non-uniform sampling, respectively. These results show that selecting the sharing side remains beneficial with a larger client population and uneven participation probabilities.}
	\begin{table}[!t]
		\centering
		\small
		\setlength{\tabcolsep}{3pt}
		\renewcommand{\arraystretch}{1.15}
		\begin{tabular*}{0.98\columnwidth}{@{\extracolsep{\fill}}lcc@{}}
			\toprule
			Method & Uniform & Non-uniform \\
			\midrule
			LoRA & 87.27\raisebox{-0.25ex}{{ $\pm$0.48}} & 83.24\raisebox{-0.25ex}{{ $\pm$0.72}} \\
			FFA-LoRA & 86.18\raisebox{-0.25ex}{{ $\pm$0.45}} & 84.90\raisebox{-0.25ex}{{ $\pm$0.85}} \\
			FedDPA-LoRA & 87.50\raisebox{-0.25ex}{{ $\pm$0.41}} & 84.29\raisebox{-0.25ex}{{ $\pm$0.64}} \\
			FedSA-LoRA & 88.12\raisebox{-0.25ex}{{ $\pm$0.55}} & 85.68\raisebox{-0.25ex}{{ $\pm$0.44}} \\
			FedAS-LoRA (Ours) & \textbf{89.37}\raisebox{-0.25ex}{{ $\pm$0.37}} & \textbf{86.80}\raisebox{-0.25ex}{{ $\pm$0.56}} \\
			\midrule
			LoRA & 93.42\raisebox{-0.25ex}{{ $\pm$0.67}} & 92.80\raisebox{-0.25ex}{{ $\pm$0.77}} \\
			FFA-LoRA & 93.61\raisebox{-0.25ex}{{ $\pm$0.71}} & 93.51\raisebox{-0.25ex}{{ $\pm$0.78}} \\
			FedDPA-LoRA & 94.87\raisebox{-0.25ex}{{ $\pm$0.52}} & 94.70\raisebox{-0.25ex}{{ $\pm$0.89}} \\
			FedSA-LoRA & 95.01\raisebox{-0.25ex}{{ $\pm$0.58}} & 96.48\raisebox{-0.25ex}{{ $\pm$0.77}} \\
			FedAS-LoRA (Ours) & \textbf{95.64}\raisebox{-0.25ex}{{ $\pm$0.73}} & \textbf{96.89}\raisebox{-0.25ex}{{ $\pm$0.79}} \\
			\midrule
			LoRA & 83.91\raisebox{-0.25ex}{{ $\pm$0.53}} & 81.25\raisebox{-0.25ex}{{ $\pm$0.60}} \\
			FFA-LoRA & 84.23\raisebox{-0.25ex}{{ $\pm$0.56}} & 83.93\raisebox{-0.25ex}{{ $\pm$0.99}} \\
			FedDPA-LoRA & 85.07\raisebox{-0.25ex}{{ $\pm$0.80}} & 83.19\raisebox{-0.25ex}{{ $\pm$0.95}} \\
			FedSA-LoRA & 86.89\raisebox{-0.25ex}{{ $\pm$0.83}} & 84.28\raisebox{-0.25ex}{{ $\pm$0.99}} \\
			FedAS-LoRA (Ours) & \textbf{87.55}\raisebox{-0.25ex}{{ $\pm$0.91}} & \textbf{85.04}\raisebox{-0.25ex}{{ $\pm$0.84}} \\
			\bottomrule
		\end{tabular*}
		\caption{Performance on the QNLI, SST-2, and MNLI-m tasks under uniform and non-uniform client sampling with $N=50$. From top to bottom, the three blocks correspond to QNLI, SST-2, and MNLI-m, respectively.}
		\label{tab:sampling_results_n50}
	\end{table}

	\subsection*{Results under Structured Input Skew}
	{\color{black}We further evaluate the compared methods under two structured input-skew partitions while controlling the client label distributions. For SST-2, we construct a label-balanced input-length-skew partition. Within each sentiment label, the samples are ordered according to their input lengths and divided into three non-overlapping groups of approximately equal size. Each client receives the corresponding length group from both sentiment labels. This construction keeps the positive and negative label proportions similar across clients, while making the client input-length distributions different.
		
		For MNLI, we construct a genre-based partition using the genre annotations provided by MultiNLI~\cite{williams2018broad}. The three clients contain premise--hypothesis pairs from the telephone, government, and fiction genres, respectively. We balance the entailment, neutral, and contradiction labels across the clients so that the main source of heterogeneity is the textual genre rather than the label distribution. We report performance on the MNLI matched evaluation set, denoted as MNLI-m.
		
		As shown in Table~\ref{tab:structured_rank_sweep}, FedAS-LoRA achieves the highest accuracy in all four evaluated settings. On SST-2, it obtains accuracies of $96.47\%$ and $96.56\%$ at $r=4$ and $r=8$, respectively, exceeding LoRA by 0.60 and 1.38 percentage points. On MNLI-m, FedAS-LoRA reaches $87.43\%$ at $r=4$ and $86.53\%$ at $r=8$, improving over LoRA by 0.80 and 0.12 percentage points. It also exceeds FFA-LoRA by 1.99 and 3.16 percentage points, and FedSA-LoRA by 0.36 and 0.34 percentage points at $r=4$ and $r=8$, respectively. These results show that selecting the sharing side remains beneficial when clients mainly differ in input characteristics rather than label proportions.}

		\begin{table}[t]
			\centering
			\small
			\renewcommand{\arraystretch}{1.15}
			
			\begin{tabular*}{0.98\columnwidth}{@{\extracolsep{\fill}}clcc@{}}
				\toprule
				Rank & Method & SST-2 & MNLI-m \\
				\midrule
				
				\multirow{4}{*}{$r=4$}
				& LoRA
				& 95.87\raisebox{-0.25ex}{{ $\pm$0.59}}
				& 86.63\raisebox{-0.25ex}{{ $\pm$0.21}} \\
				
				& FFA-LoRA
				& 95.56\raisebox{-0.25ex}{{ $\pm$0.39}}
				& 85.44\raisebox{-0.25ex}{{ $\pm$0.42}} \\
				
				& FedSA-LoRA
				& 92.98\raisebox{-0.25ex}{{ $\pm$0.45}}
				& 87.07\raisebox{-0.25ex}{{ $\pm$0.11}} \\
				
				& FedAS-LoRA (Ours)
				& \textbf{96.47}\raisebox{-0.25ex}{{ $\pm$0.19}}
				& \textbf{87.43}\raisebox{-0.25ex}{{ $\pm$0.20}} \\
				
				\midrule
				
				\multirow{4}{*}{$r=8$}
				& LoRA
				& 95.18\raisebox{-0.25ex}{{ $\pm$0.48}}
				& 86.41\raisebox{-0.25ex}{{ $\pm$0.29}} \\
				
				& FFA-LoRA
				& 93.35\raisebox{-0.25ex}{{ $\pm$0.52}}
				& 83.37\raisebox{-0.25ex}{{ $\pm$0.10}} \\
				
				& FedSA-LoRA
				& 94.04\raisebox{-0.25ex}{{ $\pm$0.44}}
				& 86.19\raisebox{-0.25ex}{{ $\pm$0.11}} \\
				
				& FedAS-LoRA (Ours)
				& \textbf{96.56}\raisebox{-0.25ex}{{ $\pm$0.04}}
				& \textbf{86.53}\raisebox{-0.25ex}{{ $\pm$0.34}} \\
				
				\bottomrule
			\end{tabular*}
			\caption{Test accuracy under structured input-skew partitions with
				different LoRA ranks $r$. SST-2 uses a label-balanced
				input-length-skew partition, while MNLI-m uses a label-balanced
				genre partition with telephone, government, and fiction clients.}
			\label{tab:structured_rank_sweep}
		\end{table}

		\subsection*{Results on GSM8K with Llama 3 8B}
		The results on the GSM8K dataset are shown in Table \ref{tab:gsm8k_qualitative_example}, demonstrating that the proposed FedAS-LoRA outperforms other methods in complex natural language generation tasks. From the given example, it can be seen that both LoRA and FFA-LoRA have reasoning errors, but FedAS-LoRA can reason accurately, demonstrating the superiority of the proposed method.
		\begin{table*}[t]
			\centering
			\small
			\setlength{\tabcolsep}{1.5pt}
			\renewcommand{\arraystretch}{1.15}
			\begin{tabular}{
					@{}
					>{\raggedright\arraybackslash}p{0.12\textwidth}
					>{\raggedright\arraybackslash}p{0.27\textwidth}
					>{\raggedright\arraybackslash}p{0.27\textwidth}
					>{\raggedright\arraybackslash}p{0.27\textwidth}
					@{}
				}
				\toprule
				& \multicolumn{1}{c}{\textbf{LoRA}}
				& \multicolumn{1}{c}{\textbf{FFA-LoRA}}
				& \multicolumn{1}{c}{\textbf{FedAS-LoRA}} \\
				\midrule
				
				\textbf{Accuracy}
				& \multicolumn{1}{c}{
					55.14\raisebox{-0.25ex}{{ $\pm$0.25}}
				}
				& \multicolumn{1}{c}{
					54.51\raisebox{-0.25ex}{{ $\pm$0.32}}
				}
				& \multicolumn{1}{c}{
					\textbf{56.28}\raisebox{-0.25ex}{{ $\pm$0.18}}
				} \\
				\midrule
				
				\textbf{Question:}
				& \multicolumn{3}{
					>{\raggedright\arraybackslash}p{0.81\textwidth}
				}{
					Every day, Wendi feeds each of her chickens three cups of mixed
					chicken feed, containing seeds, mealworms and vegetables to help
					keep them healthy. She gives the chickens their feed in three
					separate meals. In the morning, she gives her flock of chickens
					15 cups of feed. In the afternoon, she gives her chickens another
					25 cups of feed. How many cups of feed does she need to give her
					chickens in the final meal of the day if the size of Wendi's flock
					is 20 chickens?
				} \\
				\midrule
				
				\textbf{Output:}
				&
				Wendi gives her chickens 15 cups of feed in the morning and 25 cups of
				feed in the afternoon, for a total of $15 + 25 = 40$ cups of feed.
				She has 20 chickens, so each chicken needs $40 / 20 = 2$ cups of feed
				in the final meal of the day. The answer is 2.
				\textbf{[\ding{55}]}
				&
				Wendi feeds her chickens 15 cups of feed in the morning and 25 cups of
				feed in the afternoon. So she has already given them
				$15 + 25 = 40$ cups of feed. She has 20 chickens. So she needs to give
				them $40 / 20 = 2$ cups of feed each. The answer is 2.
				\textbf{[\ding{55}]}
				&
				Wendi has 15 cups of feed in the morning and 25 cups in the afternoon.
				So she has $15 + 25 = 40$ cups of feed. She has 20 chickens and she
				needs to give each chicken 3 cups of feed. So she needs
				$20 \times 3 = 60$ cups of feed. She has 40 cups of feed and she needs
				60 cups of feed. So she needs $60 - 40 = 20$ more cups of feed.
				The answer is 20. \textbf{[\ding{51}]} \\
				
				\bottomrule
			\end{tabular}
			\caption{Performance of different methods on the GSM8K dataset and
				example answers generated by each method.}
			\label{tab:gsm8k_qualitative_example}
		\end{table*}

        		\begin{table*}[htbp]
			\centering
			\small
			\setlength{\tabcolsep}{2.5pt}
			\renewcommand{\arraystretch}{1.08}
			\begin{tabular}{@{}lcccccc@{}}
				\toprule
				Method
				&  Trainable Parm.
				&  Per-round Communicated Parm.
				& \multicolumn{2}{c}{ Per-round Computation Cost}
				& \multicolumn{2}{c}{Communication Round} \\
				\cmidrule(lr){4-5}
				\cmidrule(lr){6-7}
				& (MNLI-m / SST-2)
				&
				& MNLI-m
				& SST-2
				& MNLI-m
				& SST-2 \\
				\midrule
				
				LoRA
				& 1.839M / 1.838M
				& 0.786M
				& 14.12s
				& 6.24s
				& 186
				& 210 \\
				
				FFA-LoRA
				& 1.446M / 1.445M
				& 0.393M
				& 13.94s
				& 6.04s
				& 477
				& 467 \\
				
				FedDPA-LoRA
				& 2.626M / 2.625M
				& 0.786M
				& 19.46s
				& 11.56s
				& 454
				& 398 \\
				
				FedSA-LoRA
				& 1.839M / 1.838M
				& 0.393M
				& 14.14s
				& 6.25s
				& 495
				& 453 \\
				
				FedAS-LoRA
				& 1.839M / 1.838M
				& 0.393M
				& 14.18s
				& 6.26s
				& 472
				& 384 \\
				\bottomrule
			\end{tabular}
			\caption{Time and space costs for each method on the MNLI-m and SST-2
				tasks. Communication Round denotes the communication-round index
				at which the best validation performance is obtained.}
			\label{tab:time_space_cost_mnli_sst2}
		\end{table*}
        
		\subsection*{Communication Cost}
		{\color{black}Table~\ref{tab:time_space_cost_mnli_sst2} compares the trainable parameter count, per-round communicated model size, per-round computation time, and the communication-round index of the best validation checkpoint. Share-A/Local-B and Share-B/Local-A train both LoRA factors and therefore retain the same number of trainable parameters as LoRA, i.e., 1.839M on MNLI-m and 1.838M on SST-2. However, each strategy communicates only the selected shared factor. Their per-round communicated model size is therefore 0.393M parameters, which is 50\% lower than the 0.786M parameters communicated by LoRA and FedDPA-LoRA.
			
			The reduction in per-round communication does not introduce substantial additional local computation. On MNLI-m, the per-round computation times of Share-A/Local-B and Share-B/Local-A are 14.14s and 14.18s, respectively, compared with 14.12s for LoRA. On SST-2, the corresponding times are 6.25s and 6.26s, compared with 6.24s for LoRA. FFA-LoRA has a slightly lower per-round computation time because it freezes one factor, whereas the two sharing strategies keep both factors trainable. FedDPA-LoRA incurs a higher parameter and computation cost because it maintains additional global and personalized LoRA modules. These results show that factor-wise sharing reduces the communicated model size per round while preserving the training capacity of both LoRA factors.}

		\subsection*{Relative Heterogeneity Amplification of Local LoRA Changes}
\begin{figure*}[!t]
			\centering
			
			\begin{subfigure}[t]{0.5\textwidth}
				\centering
				\includegraphics[width=\linewidth]
				{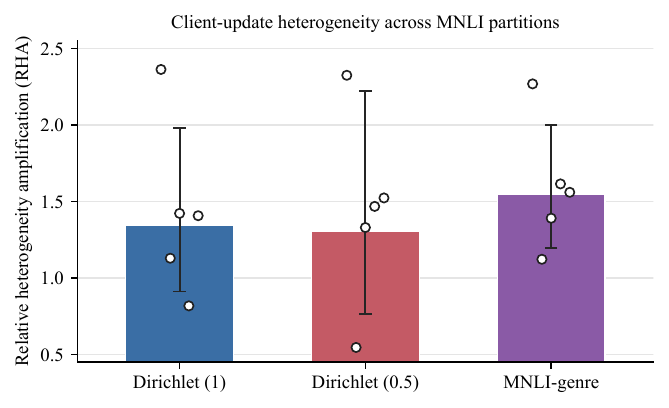}
				\caption{Overall relative heterogeneity amplification.}
				\label{fig:rha-summary}
			\end{subfigure}
			
			\begin{subfigure}[t]{0.47\textwidth}
				\centering
				\includegraphics[width=\linewidth]
				{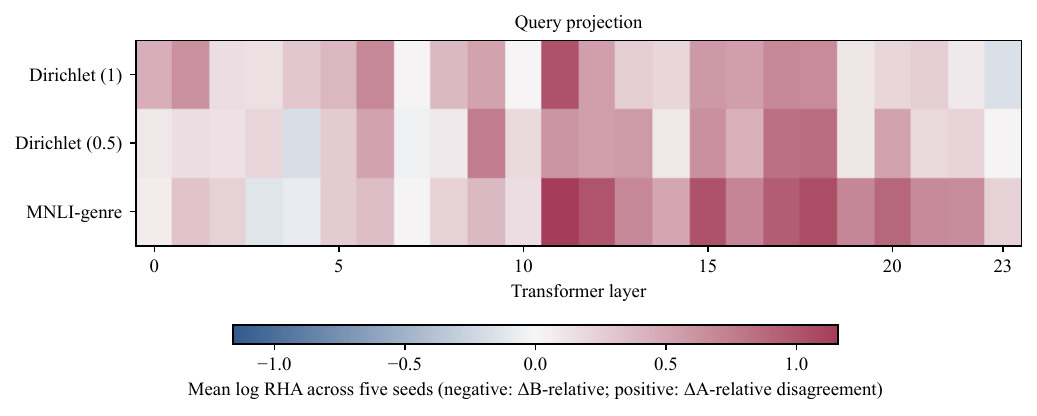}
				\caption{Layer-wise RHA for the query projection.}
				\label{fig:rha-query}
			\end{subfigure}
			\hfill
			\begin{subfigure}[t]{0.47\textwidth}
				\centering
				\includegraphics[width=\linewidth]
				{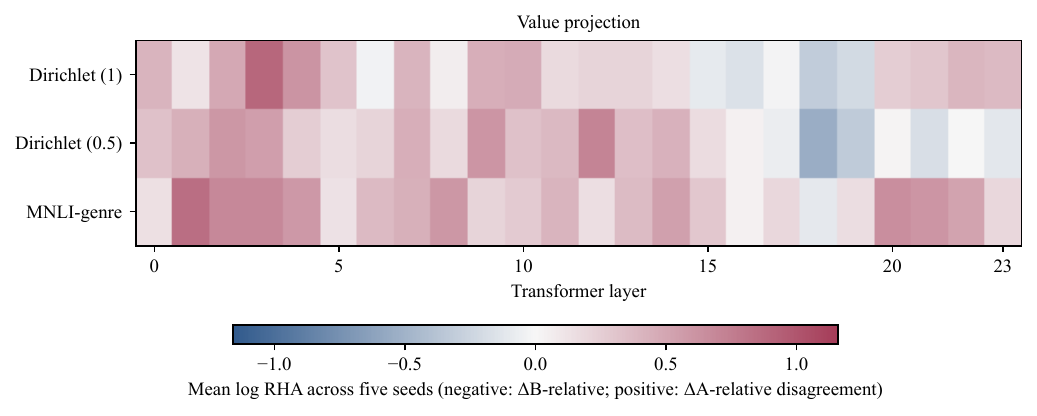}
				\caption{Layer-wise RHA for the value projection.}
				\label{fig:rha-value}
			\end{subfigure}
			
			\caption{
				Relative heterogeneity amplification of LoRA factors across
				MNLI client partitions.
				The clients perform local training independently without server
				aggregation. For each of the three non-IID partitioning schemes,
				five three-client partitions are independently generated using
				different partition seeds. Each non-IID partition is paired with an
				IID reference partition generated using the same seed, while the model
				initialization and training-related random seeds are fixed across all
				partition instances.
				(a) Bars show the geometric mean across the five partition
				instances, white circles denote individual partition instances, and
				error bars represent one sample standard deviation in log-RHA space.
				The dashed line marks $\mathrm{RHA}=1$.
				(b,c) Layer-wise mean log RHA for the query and value
				projections, respectively. Both heatmaps use the same color scale;
				positive values indicate stronger relative disagreement in
				$\Delta A$, whereas negative values indicate stronger relative
				disagreement in $\Delta B$.
			}
			\label{fig:mnli-rha}
		\end{figure*}

		{\color{black}To further examine how client data heterogeneity affects the two LoRA factors, we compare the client-specific changes of $A$ and $B$ after local training on MNLI. We consider three non-IID partitioning schemes: Dirichlet distributions with $\alpha=1$ and $\alpha=0.5$, and the label-balanced MNLI-genre partition. For each scheme, we independently generate five three-client partitions using different partition seeds. Each non-IID partition is paired with an IID partition generated using the same seed. The model initialization and all training-related random seeds are fixed across the partition instances. Therefore, the variation across the five instances is caused by the client data partitions rather than by model initialization or stochastic training.
			
			For this analysis, the clients perform local training independently without server aggregation. We measure the change in each LoRA factor relative to its common initialization as $\Delta Q_i$.
			We analyze these changes rather than the terminal factor values to exclude the initial factor values from the interpretation. This is particularly important for factor $A$, which is randomly initialized with nonzero values under the standard LoRA initialization, whereas factor $B$ is initialized to zero. Using $\Delta Q_i$ therefore places both factors on the same reference and focuses the analysis on the changes learned from each client's local data.
			
			For a given partition and factor $Q$, we quantify the cross-client disagreement using the mean pairwise Frobenius distance
			\begin{equation}
				D_Q
				=
				\frac{2}{N(N-1)}
				\sum_{1\leq i<j\leq N}
				\left\|
				\Delta Q_i-\Delta Q_j
				\right\|_F.
			\end{equation}
			For each non-IID partition instance, the relative heterogeneity amplification is defined with respect to its seed-matched IID reference as
			\begin{equation}
				\mathrm{RHA}
				=
				\frac{
					D_A^{\mathrm{non\text{-}IID}}/
					D_A^{\mathrm{IID}}
				}{
					D_B^{\mathrm{non\text{-}IID}}/
					D_B^{\mathrm{IID}}
				}.
			\end{equation}
			An RHA value greater than one means that changing from the paired IID partition to the non-IID partition produces a larger proportional increase in the cross-client disagreement of $\Delta A$ than in that of $\Delta B$. Conversely, an RHA value below one indicates a larger relative increase in the disagreement of $\Delta B$. Since RHA compares the non-IID-to-IID amplification of each factor, $\mathrm{RHA}>1$ does not necessarily imply that $D_A>D_B$ in absolute value.
			
			As shown in Fig.~\ref{fig:mnli-rha}(a), the geometric-mean RHA is above one under all three non-IID partitioning schemes. Relative to their seed-matched IID references, the evaluated non-IID partitions therefore amplify the cross-client disagreement of $\Delta A$ more strongly than that of $\Delta B$ on average. Among the three settings, the MNLI-genre partition yields the largest overall RHA, and the five individual partition instances all remain above the $\mathrm{RHA}=1$ reference line. This result indicates a relatively stable amplification of $\Delta A$ disagreement when the clients are separated by textual genre.
			
			The two Dirichlet settings show greater variation across partition instances. Although their geometric-mean RHA values are above one, each setting includes an individual instance with an RHA below one. Hence, the Dirichlet concentration parameter alone does not determine whether client heterogeneity increases the relative disagreement of $\Delta A$ or $\Delta B$ more strongly. Even under the same $\alpha$, the result depends on the realized client partition. A fixed sharing policy based only on the nominal heterogeneity level is therefore insufficient, which supports the partition-aware design of RSS.
			
			Figs.~\ref{fig:mnli-rha}(b) and~\ref{fig:mnli-rha}(c) further show that the relative amplification varies across Transformer layers and projection types. For the query projection, the mean log RHA is positive in most layers under the three partitioning schemes. The positive values are particularly evident in several middle and later layers under the MNLI-genre partition, indicating stronger relative disagreement in $\Delta A$ at these layers. The value projection exhibits a less uniform pattern. While many early and middle layers have positive mean log RHA, several later layers under the Dirichlet partitions have negative values, indicating stronger relative disagreement in $\Delta B$. These mixed layer-wise patterns show that an average factor-level trend does not fully characterize the behavior of all adapted layers and projections. This further supports evaluating shared-subspace sufficiency from the representations of the actual client partition, as done by RSS.
			
			Overall, the relative disagreement between the two LoRA factors depends on the realized client partition and varies across Transformer layers and projection types. Therefore, neither the Dirichlet concentration parameter nor an average factor-level trend is sufficient to prescribe a fixed sharing side across federated settings. RSS addresses this issue by assessing, for the current client partition and target LoRA rank $r$, whether a shared rank-$r$ input subspace is sufficient for the local data distributions. These results support selecting the sharing side according to the observed client partition rather than applying a fixed factor-sharing policy.}

\end{document}